\PassOptionsToPackage{table}{xcolor}
\documentclass{article}
\usepackage{iclr2027_conference,times}
\usepackage[T1]{fontenc} 

\usepackage{amsmath,amsfonts,bm}

\def\eqref#1{equation~\ref{#1}}

\def\1{\bm{1}}

\DeclareMathAlphabet{\mathsfit}{\encodingdefault}{\sfdefault}{m}{sl}
\SetMathAlphabet{\mathsfit}{bold}{\encodingdefault}{\sfdefault}{bx}{n}

\usepackage[hyphens]{url}
\usepackage{graphicx}
\usepackage{booktabs}
\usepackage{multirow}
\usepackage{array}
\usepackage{tabularx}
\usepackage{xcolor}
\usepackage{colortbl}
\usepackage{pifont}
\usepackage{amssymb}
\usepackage{algorithm}
\usepackage{algpseudocode}
\usepackage{capt-of}
\usepackage{fontawesome5}
\usepackage{hyperref}
\hypersetup{colorlinks=true,citecolor=black,linkcolor=black,urlcolor=black}
\colorlet{blue}{black}

\newcommand{\cmark}{\ding{51}}
\newcommand{\xmark}{\ding{55}}
\newcommand{\ptmark}{$\triangle$}
\newcommand{\nscore}{n/a}

\definecolor{tabheader}{HTML}{D6E3F0}
\definecolor{tabbest}{HTML}{C6E6C8}
\definecolor{tabours}{HTML}{E7F0D4}
\definecolor{tabalt}{HTML}{F4F6F8}

\title{SciFigQual-Bench: Scientific Figure Quality Assessment\\ with Manuscript-Linked Context}

\author{
\textbf{Zihan Deng}$^{1\dagger}$\thanks{Equal contribution. Correspondence: zhdeng@hku.hk, chuanzhi.xu@sydney.edu.au.},
\textbf{Chuanzhi Xu}$^{2\dagger}$\footnotemark[1],
\textbf{Huiqi Liang}$^{2}$,
\textbf{Xiaozhen Zhong}$^{3}$,
\textbf{Weidong Cai}$^{2}$,
\textbf{Lequan Yu}$^{1}$\\[0.45em]
$^{1}$The University of Hong Kong\\
$^{2}$The University of Sydney\\
$^{3}$University of Electronic Science and Technology of China\\[0.5em]
\href{https://frankdengai.github.io/SciFigQual-Bench/}{\faGlobe\hspace{0.35em}\textit{Project Page}}
}

\iclrfinalcopy
\makeatletter
\def\@maketitle{%
 \vbox{\hsize\textwidth
  \centering
  {\LARGE\bfseries \@title\par}
  \vskip 0.75em
  {\large \@author\par}
  \vskip 0.28in
 }%
}
\makeatother

\begin{document}
\renewcommand{\topfraction}{0.92}
\renewcommand{\bottomfraction}{0.7}
\renewcommand{\textfraction}{0.07}
\renewcommand{\floatpagefraction}{0.85}
\setcounter{topnumber}{3}
\setcounter{bottomnumber}{2}
\setcounter{totalnumber}{5}
\setlength{\textfloatsep}{8pt plus 2pt minus 2pt}
\setlength{\floatsep}{8pt plus 2pt minus 2pt}
\setlength{\intextsep}{8pt plus 2pt minus 2pt}

\maketitle
\fancyhead{}
\renewcommand{\headrulewidth}{0pt}
\pagestyle{plain}

\begin{abstract}
Scientific figure quality is bound to the manuscript: a crop can be visually clear and still contradict its caption or the paragraph that cites it.
Most image quality assessment and chart-understanding methods target a detached visual surface or a question-answering score, rather than alignment among the figure, its caption, and the citing text.
A single end-to-end prompt that sees the image and the text together still mixes what is visible with what the caption and the citing paragraph claim, so fluent wording can raise a visual score and missing text can be treated as low quality.
To address this gap, we introduce SciFigQual-Bench, a manuscript-linked benchmark for published CS-conference figures that binds each figure to its caption and to index-resolved citing paragraphs, and scores visual clarity, layout, caption consistency, context consistency, and misleading risk, leaving a dimension unevaluated when its evidence is absent.
SFQ-Agent reads the image and the text in separate calls and records modality-specific evidence.
A cross-modal judge scores caption and citing-text alignment from those reports, and a deterministic runner keeps the visual scores, sets misleading risk by a fixed rule, and applies the written caps, so each dimension follows the rubric instead of a single end-to-end prompt.
Experiments comparing this staged judge with single-pass and OCR-sidecar protocols find the closest point-estimate fit to mean human ratings under staged judging, while the gap between protocols remains small.
Caption consistency remains the main gap, and agreement with the rater mean answers a different question from agreement among human raters.
Project page: \url{https://frankdengai.github.io/SciFigQual-Bench/}
\end{abstract}

\section{Introduction}

Natural photographs are judged mainly by human visual perception: in image quality assessment (IQA), a landscape is scored on color, sharpness, and aesthetics, with no content-level right-or-wrong standard~\citep{li2025distilling}, and can often be evaluated from pixels alone~\citep{chen2024topiq}.
Scientific figures, by contrast, are bound to academic text~\citep{tufte2001,cleveland1984}: truncated axes and missing baselines are intrinsic visual-design defects readable from the canvas, whereas contradictions with captions or citing prose require manuscript context beyond pixel-only inspection.

\begin{figure}[!ht]
\centering
\includegraphics[width=0.95\linewidth]{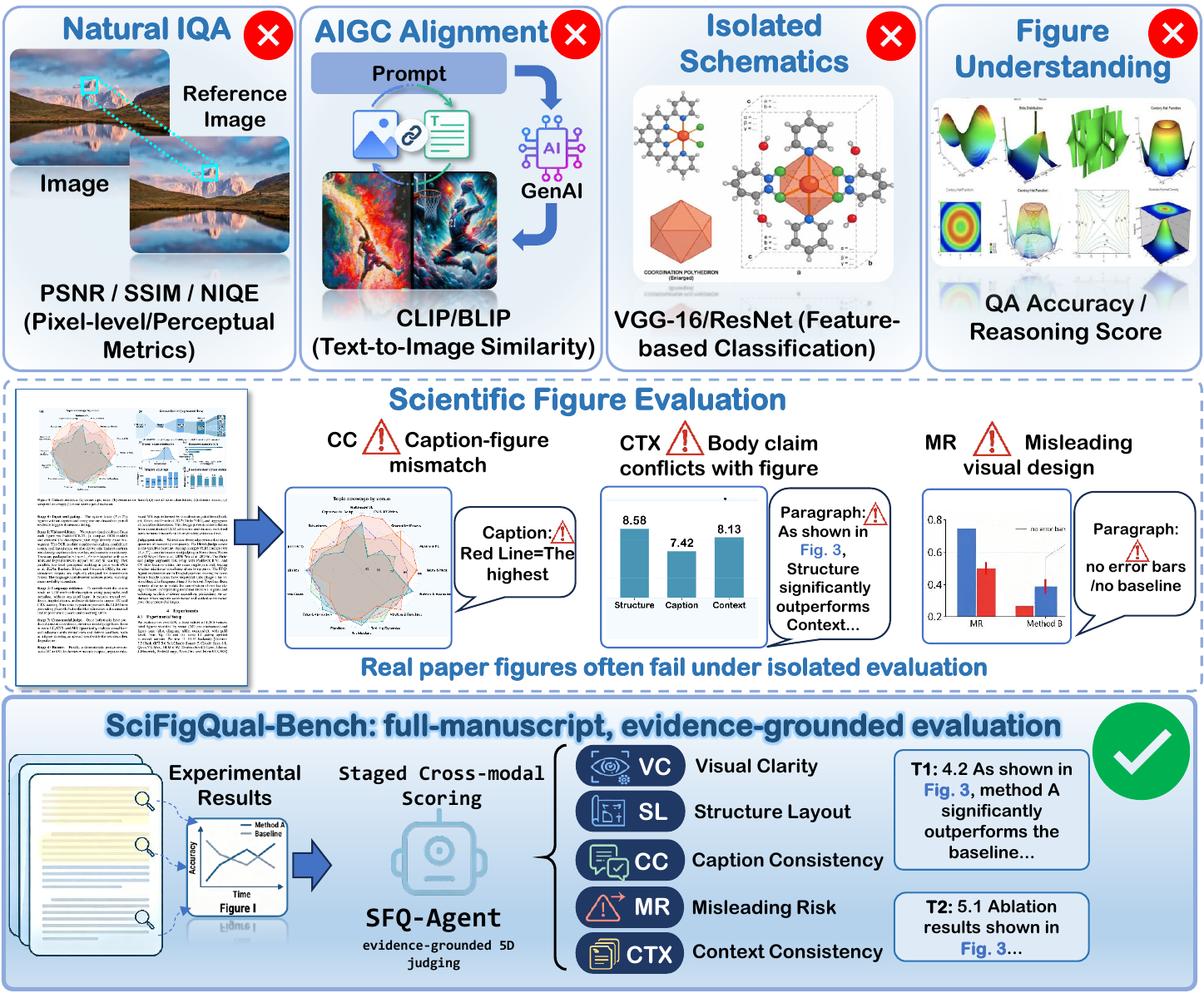}
\caption{From isolated-figure evaluation to manuscript-linked context binding. Prior paradigms evaluate figures without manuscript evidence, whereas SciFigQual-Bench grounds assessment in figure-relevant citing text and scores five separately gated dimensions on a unified 1 to 10 scale.}
\label{fig:gap}
\end{figure}

In peer review, figure quality is therefore inherently tri-modal: it depends on what is visible in the image, what the caption claims, and what the manuscript asserts in citing paragraphs.
Existing natural-IQA and AIGC-alignment paradigms operate on a single visual modality and decouple the image from both its caption and the citing manuscript.
As illustrated in Figure~\ref{fig:gap}, four mainstream paradigms highlight this gap:
Natural IQA~\citep{brisque2012,niqe2013,pieapp2018,clipiqa2022} estimates perceptual quality from low-level statistics.
AIGC alignment~\citep{radford2021clip,imagereward2023,hpsv22023,tifa2023,pickapic2023} measures text and image similarity for generated content.
Schematic understanding~\citep{chartqa2022,plotqa2020,dvqa2018,deplot2023,unichart2023} frames classification or visual question answering without tying model judgments back to the citing paragraphs of the source PDF.
Figure reasoning QA~\citep{figureseer2016,docvqa2021} tests whether a model can read a figure but does not offer a rubric for holistic quality or misleading risk.
In each case, evaluation proceeds on a detached visual input.

In this paper we propose SciFigQual-Bench for human-published scientific figures with manuscript-linked context, binding each figure to its caption and the citing-paragraph set $\mathcal{T}$ resolved from the source PDF when in-text figure references exist (otherwise $\mathcal{T}{=}\emptyset$ and CTX is withheld by {\color{blue}evidence-availability gating}, which scores a dimension only when its required evidence is present). Judge prompts receive at most two bounded citing snippets rather than the full manuscript, so our claim is contextual linkage, not long-context whole-paper reasoning.
Instead of a single end-to-end vision and language model (VLM) prompt, we collect multi-modal evidence from image, caption, and citing text, fuse it, and output traceable scores on Visual Clarity (VC), Structure \& Layout (SL), Caption Consistency (CC), Context Consistency (CTX), and Misleading Risk (MR).
Unlike faithfulness benchmarks for AI-generated scientific images~\citep{scieval2026,arxivcap2024}, the task is retrospective assessment of published figures and is complementary to generation-faithfulness evaluation. Our contributions can be summarized as follows:
\begin{itemize}
  \item We release a manuscript-linked benchmark. Published-figure quality is a judgment over the figure, its caption, and index-resolved citing paragraphs, with five dimensions and {\color{blue}evidence-availability gating}. SciFigQual-Bench provides $7{,}609$ curated figures from $1{,}144$ papers ($6{,}308$ human-rated) in top-tier CS conferences. CTX is scored only when a citing paragraph is present.
  \item We build a traceable judge. SFQ-Agent collects visual evidence and textual evidence in separate calls and combines them with a deterministic runner, so VC, SL, CC, CTX, and MR each follow the written rubric. A single-pass Direct prompt and an OCR sidecar are run on the same figures, which makes the effect of staging measurable.
  \item We measure mean fit against a rater ceiling. The frozen \textit{eval1200} split has $1{,}200$ figures, dimension-first human gold, and the same {\color{blue}evidence-availability mask} at inference. Model Within-1 is taken against that mean, and pairwise human agreement on the same split is reported as the rater ceiling.
\end{itemize}

\section{Related Work}
\label{sec:related}

Existing work that can be applied to scientific figure quality assessment and multimodal figure and text evaluation falls into four categories: perceptual IQA, chart comprehension, text-to-figure generation and multimodal corpora, and LLM-based automated evaluation.

Perceptual IQA and Isolated Image Quality.
Classical no-reference IQA (BRISQUE, NIQE) and learned perceptual metrics (PieAPP, CLIP-IQA) estimate quality from low-level statistics or deep features~\citep{brisque2012,niqe2013}.
SIQA defines a four-dimensional rubric for schematic scientific figures and trains learned regressors on expert ratings, yet its instances are not drawn from published CS PDFs and carry no index-resolved citing paragraphs~\citep{siqa2026}.
VisJudge-Bench employs MLLM judges for visualization aesthetics on chart and diagram crops~\citep{visjudge2025}, but omits source-PDF binding and manuscript-level CC/CTX.
These approaches remain visual-only on isolated crops, and we likewise do not train a specialist quality regressor: the experiments test prompted protocols on the same frozen split.

Chart and Scientific Figure Understanding.
ChartQA and SciFIBench benchmark chart reasoning and multiple-choice scientific-figure interpretation~\citep{chartqa2022,scifibench2024}.
SPUR extends to biomedical experimental images with perception, understanding, and reasoning QA and partial figure and text linkage~\citep{spur2026}, yet targets VQA accuracy rather than a calibrated multi-dimensional quality rubric and covers neither CS top-tier venues nor index-driven citing paragraphs at our scale.
These resources analyze scientific images but do not target caption and figure match, manuscript consistency, or misleading risk on official CS-conference figures.

\begin{table}[!ht]
\caption{Capability and scale comparison against related benchmarks. Symbols: \cmark=fully supported. \ptmark=partial. \xmark=not targeted.
Ours reports $7{,}609$ released figures ($6{,}308$ rated) with per-figure captions and bound citing paragraphs $\mathcal{T}$ when present (${\approx}$7.6k released text attachments).
The ${\approx}$355k citing-paragraph records and $62{,}694$ PDFs are crawl-stage construction totals, not the released inventory. Other columns follow each paper's reported test or release size.}
\label{tab:positioning}
\centering
\scriptsize
\setlength{\tabcolsep}{2.4pt}
\renewcommand{\arraystretch}{1.05}
\begin{tabular}{@{}lcccccc@{}}
\toprule
\rowcolor{tabheader}
\textbf{Comparison dimension} & \textbf{SIQA} & \textbf{SCIEval} & \textbf{VisJudge} & \textbf{SPUR} & \textbf{GENFIG1} & \cellcolor{tabours}\textbf{Ours} \\
\midrule
Real figures from published CS papers & \xmark & \xmark & \ptmark & \xmark & \xmark & \cellcolor{tabours}\cmark \\
\rowcolor{tabalt}
Source PDF + section binding & \xmark & \xmark & \xmark & \ptmark & \ptmark & \cellcolor{tabours}\cmark \\
Full-text citing paragraphs (index-driven) & \xmark & \xmark & \xmark & \ptmark & \xmark & \cellcolor{tabours}\cmark \\
\rowcolor{tabalt}
Evaluation dimensions (multi-dim.\ rubric) & \cmark & \cmark & \cmark & \ptmark & \cmark & \cellcolor{tabours}\cmark \\
End-to-end Acquire$\rightarrow$Annotate pipeline & \xmark & \ptmark & \xmark & \xmark & \ptmark & \cellcolor{tabours}\cmark \\
Auditable scoring rationale & \ptmark & \ptmark & \ptmark & \xmark & \ptmark & \cellcolor{tabours}\cmark \\
\rowcolor{tabalt}
Multi-venue $\times$ multi-year (CS top-tier) & \ptmark & \xmark & \ptmark & \xmark & \ptmark & \cellcolor{tabours}\cmark \\
\midrule
\textbf{\# Images (released)} & 2,240 & 6,000 & 3,090 & 1,084 & 584 & \cellcolor{tabours}\textbf{7,609} \\
\rowcolor{tabalt}
\textbf{\# Text items (released)} & ${\approx}$2.2k & ${\approx}$6.0k & ${\approx}$3.1k & ${\approx}$4.3k & ${\approx}$584 & \cellcolor{tabours}\textbf{${\approx}$7.6k} \\
\textbf{\# Papers (crawl scope)} & n/a & n/a & n/a & ${\approx}$5k & ${\approx}$584 & \cellcolor{tabours}\textbf{$62{,}694$} \\
\bottomrule
\end{tabular}
\end{table}

Generation Faithfulness.
SCIEval trains and benchmarks CLIP/LMM modules on multi-dimensional faithfulness of generated scientific images, covering relevance, accuracy, and interpretability in text-to-image and captioning settings~\citep{scieval2026}.
GENFIG1 asks VLMs whether automatically produced visual summaries faithfully reflect paper content~\citep{genfig12026}.
At only hundreds of instances, it targets quality control of \emph{newly synthesized} academic illustrations rather than retrospective assessment of human-authored conference figures, so extraction and captioning alone do not certify figure quality in peer-review terms.

LMM-as-Judge.
Rubric-conditioned VLM evaluators (Prometheus-Vision, VIEScore) enable fine-grained multimodal scoring~\citep{prometheus2024,viescore2024}. Evidence-grounded NLP cautions against plausible but unfaithful rationales~\citep{faithfulnlp2020}.
Monolithic judges conflate visual perception with textual verification and can hallucinate evidence when scoring CC, CTX, and MR jointly.

SciFigQual-Bench links each figure to its citing paragraphs resolved from the PDF, gathers modality-specific evidence stepwise, and fuses them only after these stages, targeting real published figures rather than isolated crops, synthetic substitutes, or multiple-choice interpretation.
Table~\ref{tab:positioning} compares the benchmark with the five closest 2024 to 2026 efforts.

\section{SciFigQual-Bench}
\label{sec:method}

\begin{figure}[!ht]
\centering
\includegraphics[width=0.8\linewidth]{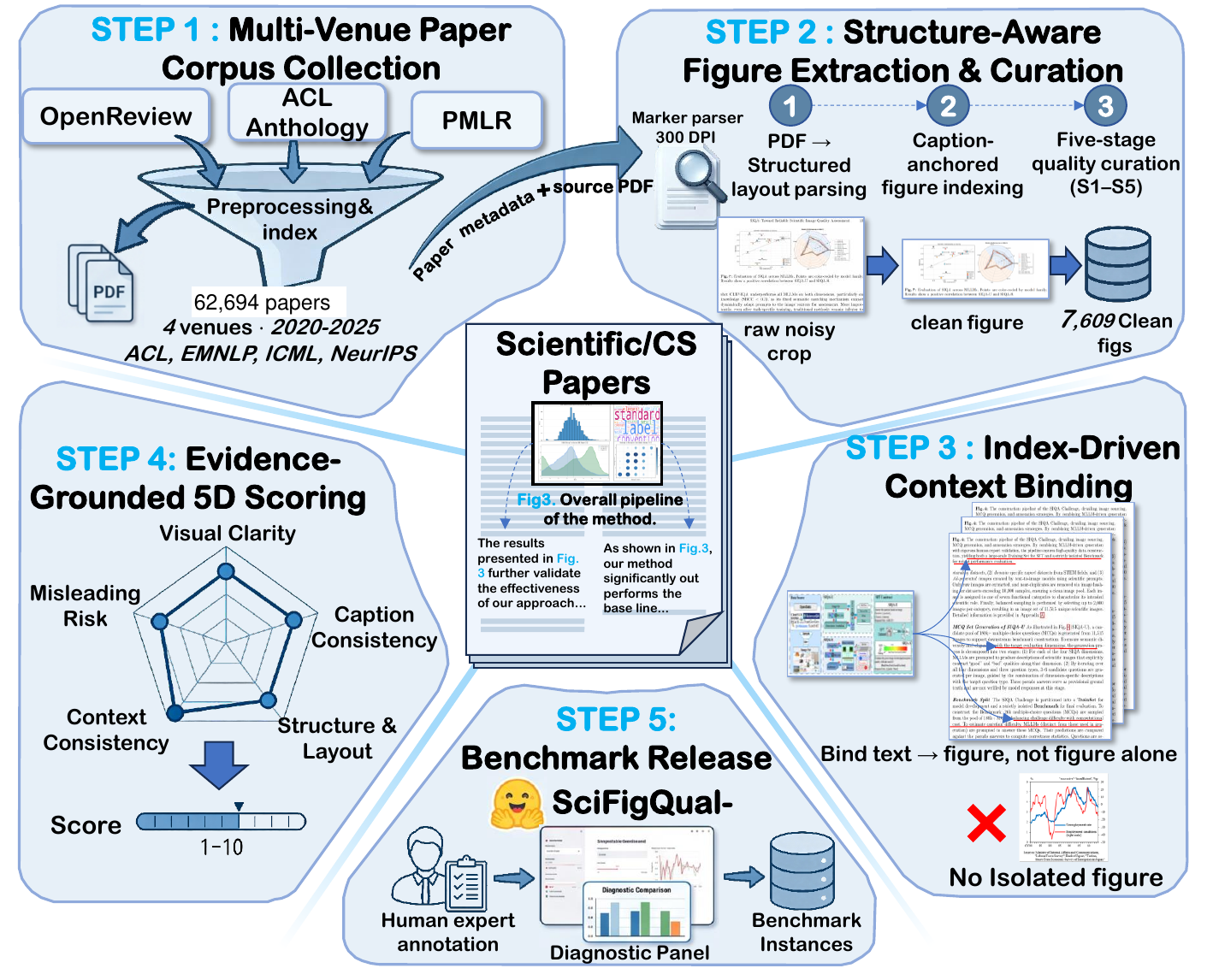}
\caption{SciFigQual-Bench construction pipeline: corpus collection, structure-aware extraction, context binding, five-dimensional rubric, and human validation.}
\label{fig:pipeline}
\end{figure}

This section defines benchmark instances and the five-dimensional rubric, then describes data construction and the SFQ-Agent scoring framework.

\subsection{Task Formulation and Rubric}
\label{sec:task}

A benchmark instance is a tuple $(I, c, \mathcal{T}, m)$ where $I$ is the figure crop, $c$ the caption (possibly empty), $\mathcal{T}$ the citing paragraphs (possibly empty), and $m$ lightweight metadata (venue, year, figure index, section tags).
Instances are always anchored to a source PDF so that $\mathcal{T}$ is resolved by figure index rather than layout heuristics alone.
Human annotators and SFQ-Agent assign scores on five separately gated dimensions, each on $[1,10]$:
Visual Clarity (VC), the legibility of text, marks, and encoding at publication scale.
Structure \& Layout (SL), composition, panel organization, and chartjunk~\citep{tufte2001}.
Caption Consistency (CC), alignment between $c$ and visible content, including subpanels and trends.
Context Consistency (CTX), alignment between evidentiary claims in $\mathcal{T}$ and what $I$ supports.
Misleading Risk (MR), the likelihood of reader misinterpretation (higher score $=$ lower risk), including truncated axes or undisclosed baselines~\citep{vlmviz2025}.
Overall quality is the mean over available dimensions and is used for protocol ranking when captions or citing text are missing.
Let $\mathcal{D}=\{\mathrm{VC},\mathrm{SL},\mathrm{CC},\mathrm{CTX},\mathrm{MR}\}$ and $g_i(d)\in\{0,1\}$ indicate whether dimension $d$ is evaluable for instance $i$ under {\color{blue}evidence-availability gating}, with human and model outputs as vectors $\mathbf{s}_i,\hat{\mathbf{s}}_i\in[1,10]^{|\mathcal{D}|}$.
The gated overall label is
\begin{equation}
y_i = \frac{1}{\sum_{d\in\mathcal{D}} g_i(d)}
\sum_{d\in\mathcal{D}} g_i(d)\,s_i(d),
\label{eq:overall}
\end{equation}
with $y_i$ undefined only when both $c$ and $\mathcal{T}$ are absent (the instance is excluded).
If $c{=}\emptyset$, CC is marked null ($g_i(\mathrm{CC}){=}0$). If $\mathcal{T}{=}\emptyset$, CTX is null.
This {\color{blue}evidence-availability gating} prevents treating missing metadata as low quality and mirrors annotation practice in SciCap+~\citep{scicapplus2023}, while MR may still be scored from visual hazards when text is partial.
{\color{blue}A caption that omits a key metric or the compared groups is capped at 6 on CC, and only after the rater judges that the omission holds for that figure. Released gold is the unclipped mean of those ratings, which Appendix~\ref{app:rubric} illustrates with Case~4.}

\subsection{Benchmark Construction}
\label{sec:construction}

Figure~\ref{fig:pipeline} summarizes the pipeline over CS papers from ACL, EMNLP, ICML, and NeurIPS (2020 to 2025).
We formalize the released benchmark as
$\mathcal{B}=\{(I_j,c_j,\mathcal{T}_j,m_j)\}_{j=1}^{N}$ with $N{=}7{,}609$ figures from $P{=}1{,}144$ qualified papers, of which $N_{\mathrm{rat}}{=}6{,}308$ carry human five-dimensional scores, while the remaining $1{,}301$ figures retain image, caption, and context bindings for protocol development and future labeling but are excluded from gold-based evaluation on \textit{eval1200}.
Each module below is deterministic given source PDFs and fixed preprocessing seeds.

\noindent \textbf{Step 1: Corpus Acquisition.}
{\color{blue}We collect PDFs and bibliographic metadata from OpenReview, ACL Anthology, and PMLR, and tag each file with $(\mathit{venue},\mathit{year},\mathit{paper\_id})$. Records outside 2020 to 2025 or missing parseable metadata are discarded. PDF linearization, publisher-watermark removal, and exact-duplicate removal are reported in Appendix~\ref{app:acquisition}.}
The surviving raw corpus is
\begin{equation}
\mathcal{C}_0=\{p_k\}_{k=1}^{K_0},\quad K_0=62{,}694,
\label{eq:corpus}
\end{equation}
larger in crawl scope than DocFigure and SciFIBench sources~\citep{docfigure2019,scifibench2024,arxivcap2024}, but used here for quality evaluation rather than MCQ or type classification.

\begin{figure}[!ht]
\centering
\includegraphics[width=\linewidth]{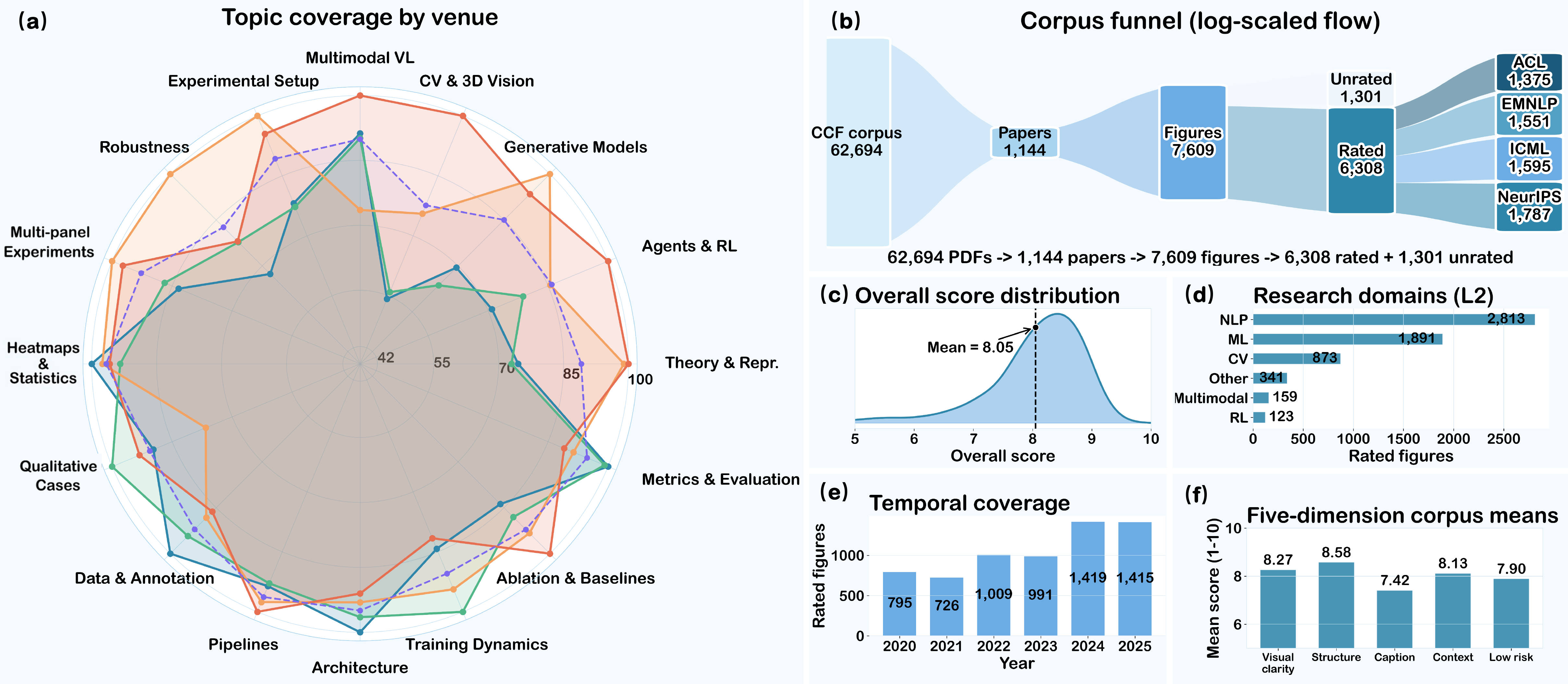}
\caption{Corpus overview of SciFigQual-Bench. From $62{,}694$ PDFs we retain $1{,}144$ papers and $7{,}609$ figures ($6{,}308$ human-rated) across ACL/EMNLP/ICML/NeurIPS (2020 to 2025). Overall scores concentrate near $8.05$. Caption consistency is the weakest axis.}
\label{fig:stats}
\end{figure}

\noindent \textbf{Step 2: Structure-Aware Figure Extraction.}
{\color{blue}Each retained PDF is rasterized and parsed into text blocks, figure boxes, and caption candidates. Candidates are ranked by spatial overlap, vertical distance, and a figure-caption cue~\citep{figex2025}. The rasterization resolution and the layout parser are given in Appendix~\ref{app:acquisition}.}
A five-stage curation pipeline then removes duplicates and non-figures, filters anomalous crops, recovers clipped panels, flags borderline cases, and discards irrecoverable layouts (Appendix~\ref{app:curation}).

\noindent \textbf{Step 3: Index-Driven Context Binding.}
We detect figure references in the body with PyMuPDF by matching index patterns, collect the hosting paragraph for each match, and merge multi-sentence windows within the same section, yielding per-figure context sets $\mathcal{T}_k = \bigcup_{\tau\in\text{match}(k)} u(\tau)$, each capped at $L_{\max}$ tokens.
Index matching over the raw crawl yields approximately 355,000 citing-paragraph records. These are index hits rather than abstract or introduction heuristics~\citep{s1mmalign2026,arxivcap2024}.
After curation, each of the $7{,}609$ released figures is paired with its caption and the resolved $\mathcal{T}$ (possibly empty). Scoring prompts use at most two snippets (Appendix~\ref{app:context}).
{\color{blue}Annotators read those index-resolved paragraphs in full, and the paragraphs average 748 characters, which makes the two-snippet model input a conservative comparison.}
Paragraphs lacking explicit figure-specific claims are down-weighted, and comparative or trend language is preferred for subsequent CTX annotation.

\noindent \textbf{Step 4: Five-Dimensional Human Annotation.}
Trained human annotators were calibrated on a 50-figure pilot using the rubric in Appendix~\ref{app:rubric}, after which each instance received scores under Eq.~\ref{eq:overall} with {\color{blue}evidence-availability gating}, so that CC and CTX were hidden whenever the necessary evidence was absent.
Following faithful-NLP guidelines~\citep{faithfulnlp2020}, annotators had to justify every score with specific visible marks, caption phrases, or citing sentences, and any rationale lacking evidence was returned for revision.
We performed dual annotation on an ACL 2025 holdout, with disagreements exceeding two points triggering adjudication.
{\color{blue}Sixteen annotators scored \textit{eval1200}, and each rated between 219 and 264 figures.}
The released gold for \textit{eval1200} uses at least three raters per figure, aggregated dimension-first (Appendix~\ref{app:aggregation}). On complete ratings this coincides with the mean of per-rater overalls, and the two can diverge only under uneven {\color{blue}evidence-availability null} masks across raters. The final set comprises $6{,}308$ rated figures.

\noindent \textbf{Step 5: Release and Evaluation Packaging.}
We package all rated records, context indices $\mathcal{T}$, and model-score outputs in a unified JSONL schema, together with a frozen evaluation split, \textit{eval1200}, of 1,200 paper-aware instances mixed by figure type. Gold labels were frozen before model calls. The split and its gold are released for reproducibility after protocol freeze and should not be treated as a permanently hidden leaderboard test.
Three judge protocols (Direct, Sidecar, and SFQ-Agent) share the same input fields $(I,c,\mathcal{T},m)$ and the same five-dimensional schema, and differ only in evidence staging and call budget.

\subsection{Dataset Statistics}
\label{sec:stats}

Figure~\ref{fig:stats} summarizes scale, diversity, and quality relative to prior scientific-figure benchmarks.
Radar coverage in (a) spans 18 research categories across four venues, broader than chart-only sets such as CharXiv~\citep{charxiv2024}.
The funnel in (b) yields $P{=}1{,}144$ papers and $N{=}7{,}609$ figures from $K_0{=}62{,}694$ PDFs (Eq.~\ref{eq:corpus}), of which $6{,}308$ are human-rated, exceeding SIQA-U (2,240) and SPUR (1,084) in comparable settings~\citep{siqa2026,spur2026}. Paper retention is $P/K_0\approx 1.8\%$, and figure yield after curation is $N/K_0\approx 0.12$ figures per crawled PDF.
Overall scores in (c) concentrate near 8.05 because published figures are mostly high quality, so ranking is harder than coarse Within-1.
Rated figures in (d) are dominated by NLP (2,813), ML (1,891), and CV (873), with the remainder spanning other CS areas, while temporal coverage in (e) peaks in 2024 to 2025 with over 1,400 rated figures per year.
Mean dimension scores in (f) show CC (7.42) as the weakest axis versus SL (8.58): caption and figure mismatches, rather than blur, are the typical defects in this published-only pool.

\subsection{SFQ-Agent}
\label{sec:agent}

\begin{figure}[!ht]
\centering
\includegraphics[width=\linewidth]{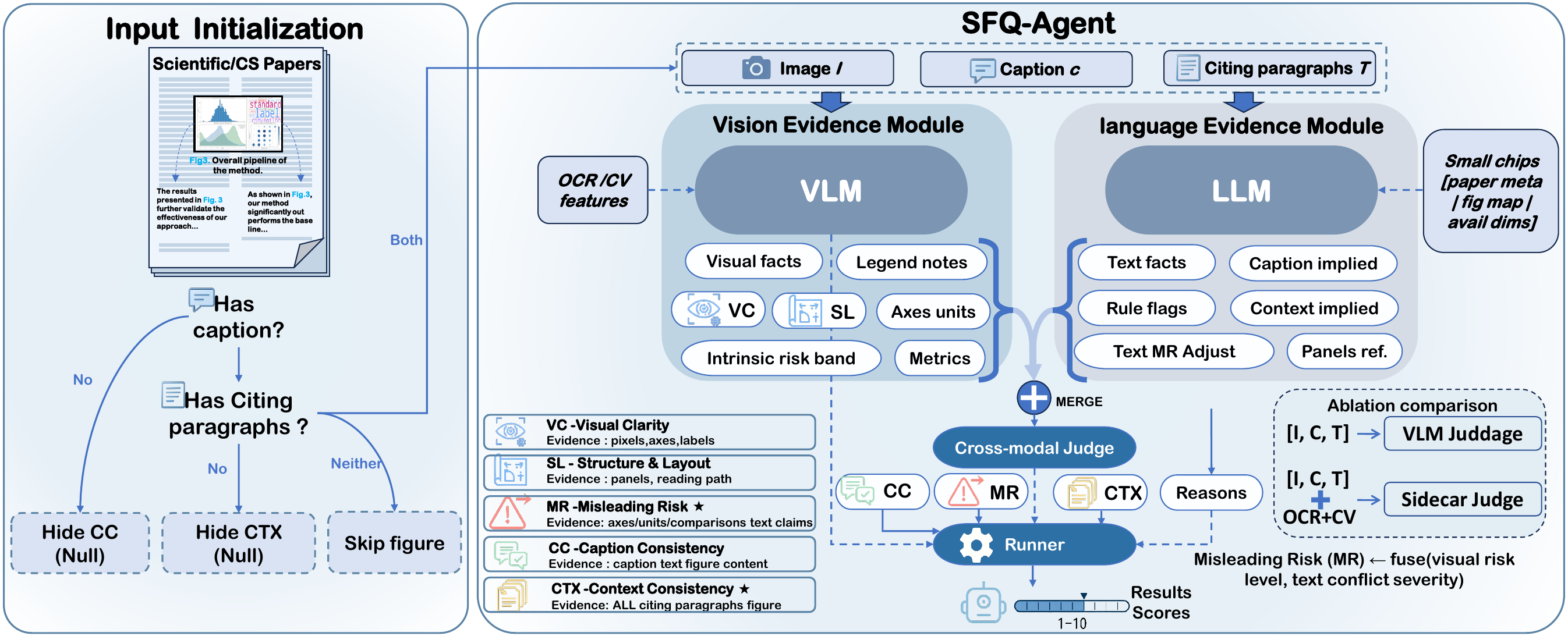}
\caption{SFQ-Agent scoring pipeline: {\color{blue}evidence-availability gating}, parallel vision and language evidence modules, cross-modal judge, and deterministic Runner aggregation. Direct and Sidecar judges serve as ablations.}
\label{fig:agent}
\end{figure}

The statistics above identify caption and context consistency as the main bottleneck, which a single end-to-end VLM struggles to score.
SFQ-Agent (Figure~\ref{fig:agent}) implements evidence-grounded judging as an auditable alternative to monolithic LMM-as-judge pipelines~\citep{llm_judge2023,prometheus2024,viescore2024}.

\noindent \textbf{Step 0: Input and Gating.}
The system loads $(I, c, \mathcal{T})$, discards figures that lack both a caption and citing text, and hides dimensions when evidence is only partial.
\noindent \textbf{Step 1: Vision Evidence.}
We extract visual evidence via PaddleOCR-VL and classical CV features (blur, edge density, color histogram).
The OCR/CV module supplies text regions, confidence, and layout cues for VC/SL, packaged with axis/legend notes as visual facts. Precomputed caption and OCR token/numeric overlap ratios are side features for later CC fusion, not substitutes for pixel-level VC/SL.
The language module never accesses pixels.
\noindent \textbf{Step 2: Language Evidence.}
An LLM reads only the caption, citing paragraphs, and metadata, with no pixel input, and extracts textual evidence, implied claims, and text-side severity/caps (omissions, bare references) to support later CC/CTX/MR fusion.
This separation prevents plausible but unfaithful rationales~\citep{faithfulnlp2020}.
{\color{blue}Appendix~\ref{app:faithfulness} checks that claim on 100 figures from \textit{eval1200}.}
\noindent \textbf{Step 3: Cross-Modal Judge.}
The judge fuses the two evidence reports to score CC/CTX and to write an MR rationale, aligning visual hazard indicators with textual cues and detecting conflicts (e.g., an upward trend in the figure versus ``degradation'' in the text).
VC and SL are copied from Stage~1 rather than re-estimated from text summaries, so fluent language cannot re-inflate visual scores.
\noindent \textbf{Step 4: Runner.}
A deterministic post-processor locks VC and SL to the vision-module outputs, replaces the MR number by the closed-form fusion of $[\ell,h]$ and $\sigma$ (Algorithm~\ref{alg:runner}).
{\color{blue}Appendix~\ref{app:model-cases} walks through those quantities on Ex-H.}
It also applies CC/CTX caps, and aggregates under {\color{blue}evidence-availability gating}, which prevents score inflation from unconstrained LLM arbitration and keeps each final score traceable to its originating evidence track.
{\color{blue}Prompt wording remains in Appendix~\ref{app:prompts}. The call contract and the fusion rule are stated here.}

{\color{blue}
\noindent \textbf{Why the vision call and the language call are separate.}
One call puts the caption and the citing text beside the image, and fluent text can lift the visual scores.
Table~\ref{tab:protocol-ablation} records that staging reduces this lenient bias, from $+0.022$ to $+0.012$ for GPT-5.6-Sol and from $+0.102$ to $+0.072$ for GLM-4.6V.
CC and CTX are a comparison, so the two descriptions have to be written independently.
The vision call states what the image shows and what a faithful caption and citing paragraph should say.
The language call states what the caption and the citing text imply that the image shows, and it does not see the image.
This is the same idea as factored verification in Chain-of-Verification~\citep{dhuliawala2024chain}, where a check is answered independently of the draft it checks.
The split also fixes ownership. VC and SL come only from the vision call, the caps and the {\color{blue}text-claim severity} come only from the language call, and Algorithm~\ref{alg:runner} is code, so the same prompts determine the same numbers.

\noindent \textbf{What each call may write.}
The vision call reads the image and the OCR/CV features, writes the final VC and SL, and emits image facts with an intrinsic risk band $[\ell,h]$.
It does not score CC, CTX, or MR.
The language call reads only the caption, the citing text, and metadata.
It emits omission and bare-reference caps, usually $6$, and a {\color{blue}text-claim severity} $\sigma\in\{\mathrm{none},\mathrm{minor},\mathrm{moderate},\mathrm{severe},\mathrm{contradiction}\}$ from the wording alone, with no final scores and no missing-mark judgment.
The fusion call reads the two reports, proposes CC and CTX, and writes an MR rationale.
Any MR number it emits is a draft.
Algorithm~\ref{alg:runner} copies VC and SL from the vision call, applies the caps, clamps CTX to at most $2$ under a contradiction, and replaces the draft MR.
Without that post-processor, staged judging is not reproducible from the prompts alone.
{\color{blue}A conflict visible only when the two reports are compared is created by fusion and written into CC and CTX.}
MR continues to use the vision band $[\ell,h]$ and the text-claim severity $\sigma$.
When $\sigma$ is contradiction, Algorithm~\ref{alg:runner} sets MR to $2$ and discards the fusion draft.
{\color{blue}Appendix~\ref{app:model-cases} walks through Ex-H under this split and separates the closed-form MR demonstration from the stored F3 value in Figure~\ref{fig:apx-human-model}.}

\begin{algorithm}[!ht]
\color{blue}
\footnotesize
\caption{SFQ-Agent Runner: ownership, caps, and MR fusion.}
\label{alg:runner}
\begin{algorithmic}[1]
\Require scores $s$. Band $[\ell,h]$. Severity $\sigma$. Caps $\mathrm{cap}_{\mathrm{CC}}$, $\mathrm{cap}_{\mathrm{CTX}}$, and an optional MR cap from Stage~2
\State $s(\mathrm{VC})\gets s_{\mathrm{VLM}}(\mathrm{VC})$. $s(\mathrm{SL})\gets s_{\mathrm{VLM}}(\mathrm{SL})$
\State \textbf{if} CC on \textbf{and} $\mathrm{cap}_{\mathrm{CC}}$ set \textbf{then} $s(\mathrm{CC})\gets\min(s(\mathrm{CC}),\mathrm{cap}_{\mathrm{CC}})$
\State \textbf{if} CTX on \textbf{and} $\mathrm{cap}_{\mathrm{CTX}}$ set \textbf{then} $s(\mathrm{CTX})\gets\min(s(\mathrm{CTX}),\mathrm{cap}_{\mathrm{CTX}})$
\State \textbf{if} CTX on \textbf{and} $\sigma{=}\mathrm{contradiction}$ \textbf{then} $s(\mathrm{CTX})\gets\min(s(\mathrm{CTX}),2)$
\State $m\gets(\ell+h)/2$
\State $r\gets\begin{cases}
\mathrm{none}:\ m\\
\mathrm{minor}:\ \max(\ell,m{-}1)\\
\mathrm{moderate}:\ \min(m,6.5)\\
\mathrm{severe}:\ \min(m,4.5)\\
\mathrm{contradiction}:\ 2
\end{cases}$
\State $s(\mathrm{MR})\gets\mathrm{clip}_{[1,10]}(\mathrm{round}(r))$
\State \textbf{if} $\mathrm{cap}_{\mathrm{MR}}$ set \textbf{then} $s(\mathrm{MR})\gets\min(s(\mathrm{MR}),\mathrm{cap}_{\mathrm{MR}})$
\State \Return $s$ and {\color{blue}evidence-availability-gated} mean over scored dims
\end{algorithmic}
\end{algorithm}
}

Judge Protocols.
We evaluate three protocols of increasing complexity.
The Direct Judge issues a single VLM prompt over $(I, c, \mathcal{T})$, similar to zero-shot judging in Prometheus-Vision and Q-Align~\citep{prometheus2024,qalign2024}. The Sidecar Judge adds PaddleOCR-VL and CV side features within the same single-pass call. SFQ-Agent reuses the same feature bundle across three sequential calls (vision, language, fusion).
The variants isolate OCR-VL side signals versus explicit evidence staging, especially on CC and CTX, where monolithic judges most often conflate perception with text verification.

\section{Experiments and Results}
\label{sec:experiments}

\subsection{Experimental Setup}

We evaluate on \textit{eval1200}: 1,200 figures from 254 papers, sampled paper-wise (ACL 266, EMNLP 294, ICML 294, NeurIPS 346) and mixed by figure type (plot, diagram, table, composite). Gold follows Eq.~\ref{eq:overall}, and model outputs use the same {\color{blue}evidence-availability mask} (Appendix~\ref{app:eval1200}).
CTX is scored on the 807 figures with citing text. Three of them keep only one CTX rating after per-rater nulls, so Table~\ref{tab:human-ceiling} uses the remaining $n{=}804$.
Eleven backends run under Direct and, where available, Sidecar. SFQ-Agent adds Gemini-3.1-Pro and matches vision with language except for F5 (Qwen-VL-Max vision with Qwen-Plus language). The 29 rows share $(I,c,\mathcal{T},m)$ and do not fill an $11{\times}3$ grid, because Sidecar omits the Claude backends (Appendix~\ref{app:runids}).
Direct is one VLM call that returns every evaluable dimension together. Sidecar keeps that single call and adds PaddleOCR-VL and CV features, testing whether extra visual cues help without changing the decision. SFQ-Agent uses three calls (Section~\ref{sec:agent}): vision locks VC/SL, language reads caption and citing text without pixels, and fusion scores CC, CTX, and MR, after which the Runner locks the visual scores and applies the caps. The comparison therefore separates denser OCR observation from explicit modality separation (Appendix~\ref{app:prompts}).
Metrics on the full split, unless a table says otherwise, are MAE, Within-1 (within $\pm1$), Spearman correlation, and signed bias (positive $=$ lenient) versus mean gold. CC, CTX, and MR additionally report MAE on {\color{blue}evidence-available} items only. A constant predictor of the split mean $8.13$ is the floor (MAE $0.565$, W-1 $87.9\%$).

\subsection{Protocol Ablation Results}
\label{sec:ablation}

Table~\ref{tab:protocol-ablation} consolidates the main evaluation on \textit{eval1200}: 29 protocol and backend configurations spanning Direct, Sidecar, and SFQ-Agent.
Rows share $(I,c,\mathcal{T})$ and {\color{blue}evidence-availability gating}, but Agent uses three calls per figure and Sidecar is missing Claude backends, so rows are comparable on inputs rather than on compute.

\begin{table}[!ht]
\caption{Protocol and backend ablation on \textit{eval1200} ($n{=}1200$) versus \emph{mean} gold.
$\uparrow$ better. $\downarrow$ worse.
W-1 is not pairwise human agreement.
Green: best per protocol (lowest MAE). Sidecar ``+OCR'' uses PaddleOCR-VL.}
\label{tab:protocol-ablation}
\centering
\scriptsize
\renewcommand{\arraystretch}{0.95}
\begin{tabularx}{0.9\linewidth}{@{}Xccccccc@{}}
\toprule
\rowcolor{tabheader}
\textbf{Backend} &
\textbf{MAE}$\downarrow$ & \textbf{W-1}$\uparrow$ & \textbf{SRCC}$\uparrow$ & \textbf{Bias} &
\textbf{CC}$\downarrow$ & \textbf{CTX}$\downarrow$ & \textbf{MR}$\downarrow$ \\
\midrule
\multicolumn{8}{@{}l}{\textbf{Direct} (1 call)} \\
Gemini-3.5-Flash       & 0.464 & 91.0\% & 0.532 & 0.032 & 0.942 & 0.972 & 0.724 \\
GPT-5.6-Sol         & 0.448 & 92.0\% & 0.565 & 0.022 & 0.922 & 0.928 & 0.708 \\
Claude-Sonnet-5       & 0.478 & 90.2\% & 0.518 & 0.038 & 0.968 & 0.862 & 0.732 \\
Qwen-VL-Max         & 0.520 & 88.8\% & 0.416 & 0.065 & 0.985 & 1.044 & 0.758 \\
GLM-4.6V           & 0.662 & 81.0\% & 0.438 & 0.102 & 1.082 & 1.128 & 0.812 \\
Doubao-Seed-2.0-pro     & 0.609 & 83.2\% & 0.570 & -0.120 & 0.878 & 0.918 & 0.752 \\
Llama-4-Maverick       & 0.486 & 89.8\% & 0.512 & 0.035 & 0.958 & 0.948 & 0.728 \\
Pixtral-Large        & 0.502 & 89.2\% & 0.498 & 0.042 & 0.972 & 0.968 & 0.742 \\
Nova-Pro           & 0.471 & 90.6\% & 0.528 & 0.028 & 0.938 & 0.912 & 0.716 \\
\rowcolor{tabbest}
\textbf{Claude-Opus-4.8}   & \textbf{0.443} & \textbf{92.4\%} & \textbf{0.582} & \textbf{0.015} & \textbf{0.908} & \textbf{0.888} & \textbf{0.692} \\
InternVL3-78B        & 0.538 & 87.6\% & 0.445 & 0.058 & 1.012 & 1.052 & 0.768 \\
\midrule
\multicolumn{8}{@{}l}{\textbf{Sidecar} (1 call + OCR)} \\
Gemini-3.5-Flash + OCR    & 0.452 & 91.8\% & 0.546 & 0.026 & 0.928 & 0.958 & 0.712 \\
\rowcolor{tabbest}
\textbf{GPT-5.6-Sol + OCR}  & \textbf{0.436} & \textbf{92.6\%} & \textbf{0.578} & \textbf{0.018} & \textbf{0.912} & \textbf{0.918} & \textbf{0.698} \\
Qwen-VL-Max + OCR      & 0.516 & 89.1\% & 0.470 & 0.050 & 0.972 & 0.962 & 0.778 \\
GLM-4.6V + OCR        & 0.618 & 83.8\% & 0.472 & 0.088 & 1.048 & 1.088 & 0.792 \\
Doubao-Seed-2.0-pro + OCR   & 0.596 & 84.6\% & 0.584 & -0.098 & 0.862 & 0.898 & 0.738 \\
Nova-Pro + OCR        & 0.459 & 91.2\% & 0.538 & 0.022 & 0.928 & 0.902 & 0.708 \\
Llama-4-Maverick + OCR     & 0.474 & 90.4\% & 0.524 & 0.030 & 0.948 & 0.938 & 0.718 \\
Pixtral-Large + OCR     & 0.490 & 89.6\% & 0.508 & 0.036 & 0.962 & 0.958 & 0.732 \\
InternVL3-78B + OCR     & 0.524 & 88.2\% & 0.458 & 0.052 & 0.998 & 1.038 & 0.758 \\
\midrule
\multicolumn{8}{@{}l}{\textbf{SFQ-Agent} (3 calls)} \\
Gemini-3.5-Flash       & 0.440 & 92.6\% & 0.560 & 0.020 & 0.914 & 0.942 & 0.698 \\
Gemini-3.1-Pro        & 0.428 & 92.9\% & 0.572 & 0.016 & 0.906 & 0.932 & 0.692 \\
\rowcolor{tabbest}
\textbf{GPT-5.6-Sol}     & \textbf{0.418} & \textbf{93.4\%} & \textbf{0.598} & \textbf{0.012} & \textbf{0.892} & \textbf{0.896} & \textbf{0.678} \\
Claude-Sonnet-5       & 0.456 & 91.2\% & 0.542 & 0.028 & 0.948 & 0.842 & 0.718 \\
Qwen-VL-Max+Qwen-Plus    & 0.511 & 89.3\% & 0.519 & 0.030 & 1.007 & 0.833 & 0.938 \\
GLM-4.6V           & 0.574 & 86.4\% & 0.508 & 0.072 & 1.012 & 1.052 & 0.768 \\
Llama-4-Maverick       & 0.462 & 91.0\% & 0.532 & 0.022 & 0.942 & 0.932 & 0.712 \\
Claude-Opus-4.8       & 0.424 & 93.1\% & 0.588 & 0.014 & 0.898 & 0.882 & 0.686 \\
Nova-Pro           & 0.446 & 92.2\% & 0.548 & 0.018 & 0.924 & 0.906 & 0.702 \\
\bottomrule
\end{tabularx}
\end{table}

Direct Baseline.
Claude-Opus-4.8 leads Direct protocols (MAE 0.443, W-1 92.4\%, SRCC 0.582), closely followed by GPT-5.6-Sol.
Qwen-VL-Max is reasonably calibrated (W-1 88.8\%) but ranks poorly (SRCC 0.416), driven by high CC/CTX errors ($\mathrm{MAE}_{\mathrm{CC}}\approx0.99$, $\mathrm{MAE}_{\mathrm{CTX}}\approx1.04$), while Doubao-Seed-2.0-pro shows strong negative bias ($-0.120$), indicating systematic harshness, and GLM-4.6V lags with the largest errors across dimensions.
Sidecar Ablation.
On every backend that has a Sidecar row, adding PaddleOCR-VL and CV features reduces MAE, with the largest drop on GLM ($\Delta$MAE$=0.044$) and modest CC/CTX improvements for GPT-5.6-Sol (CC $0.922{\rightarrow}0.912$, CTX $0.928{\rightarrow}0.918$).
Sidecar is not run for Claude backends, so the claim holds only over the nine Sidecar rows.
SFQ-Agent Ablation.
On matched backends, staged judging lowers overall MAE relative to Direct, at three calls per figure rather than one.
GPT-5.6-Sol has the lowest observed overall error in the frozen matrix (MAE 0.418, W-1 93.4\%, SRCC 0.598, bias $+0.012$) and the lowest observed MR error (0.678), above the constant-mean floor (MAE 0.565, W-1 87.9\%), for a Direct-best to Agent-best point-estimate gap of $0.025$ MAE. We do not attach a significance claim to this gap.
Among matched Agent runs, Claude-Sonnet-5 attains the lowest CTX MAE (0.842). The heterogeneous F5 run has a still lower CTX MAE (0.833) but raises MR MAE to 0.938 (Direct Qwen 0.758), so that pairing does not by itself resolve visual and text conflicts.
{\color{blue}Outside Table~\ref{tab:protocol-ablation}, we add a same-backbone comparison of ReAct and SFQ-Agent on 120 figures with local images, 30 from each of ACL, EMNLP, ICML, and NeurIPS.
Compute and time limits in the rebuttal window restrict this release to that subset, and the same comparison will be extended to the full 1,200 figures in a follow-up.
Overall MAE falls from 1.461 to 0.873 on \texttt{doubao-seed-2-1-pro-260628} and from 1.493 to 0.746 on \texttt{gpt-5.6-sol}.
These values are for this 120-figure ReAct comparison. They do not replace the \textit{eval1200} MAE in Table~\ref{tab:protocol-ablation}.
The Doubao backbone here is Doubao-Seed-2.1-Pro, separate from Doubao-Seed-2.0-pro in that table.
SFQ-Agent is lower on every reported dimension.
ReAct may revise visual clarity and structure layout after the caption.
SFQ-Agent locks those two scores on the vision call and fuses misleading risk in the Runner.
Both protocols use temperature 0, disabled extended reasoning, and the paper prompts, with no OCR or CV features.
Mean calls logged on 40 of these figures are 3.00 for SFQ-Agent, and 3.73 and 3.40 for ReAct on the two backbones.
Appendix Table~\ref{tab:apx-react-sfq} gives the full matrix and the logged token counts.}

\subsection{Human Ceiling}
\label{sec:ceiling}

\noindent
\begin{minipage}[t]{0.48\linewidth}
\vspace{0pt}
Table~\ref{tab:protocol-ablation} compares models to \emph{mean} gold, whereas Table~\ref{tab:human-ceiling} reports the rater ceiling on the same split and does not restate model MAE: human MAE is leave-one-out versus the other raters, W-1 is pairwise, and $\alpha$ is ordinal Krippendorff.
Overall human pairwise W-1 is 71.5\% ($\alpha{=}0.51$), so Agent W-1 $93.4\%$ in Table~\ref{tab:protocol-ablation} is versus mean gold, not versus a rater.
CC is the unstable construct (pairwise W-1 52.5\%, $\alpha{=}0.49$, LOO MAE 1.50), and MAE/W-1 versus the mean therefore cannot be read as superhuman caption verification.
\end{minipage}\hspace{0.04\linewidth}%
\begin{minipage}[t]{0.43\linewidth}
\vspace{0pt}
\captionof{table}{Human agreement ceiling on \textit{eval1200} (no model scores).
MAE: leave-one-out. W-1: pairwise. $\alpha$: ordinal Krippendorff.
CTX: $n{=}804$ with $\ge$2 ratings.}
\label{tab:human-ceiling}
\centering
\scriptsize
\renewcommand{\arraystretch}{1.08}
\begin{tabularx}{\linewidth}{@{}Xccc@{}}
\toprule
\rowcolor{tabheader}
\textbf{Dim.} & \textbf{MAE}$\downarrow$ & \textbf{W-1}$\uparrow$ & \textbf{$\alpha$}$\uparrow$ \\
\midrule
VC & 0.87 & 76.6\% & 0.44 \\
\rowcolor{tabalt}
SL & 0.62 & 84.8\% & 0.42 \\
CC & 1.50 & 52.5\% & 0.49 \\
\rowcolor{tabalt}
CTX & 0.88 & 77.7\% & 0.61 \\
MR & 1.12 & 67.9\% & 0.45 \\
\midrule
Overall & 0.70 & 71.5\% & 0.51 \\
\bottomrule
\end{tabularx}
\end{minipage}

\begin{figure}[!ht]
\centering
\includegraphics[width=0.99\linewidth]{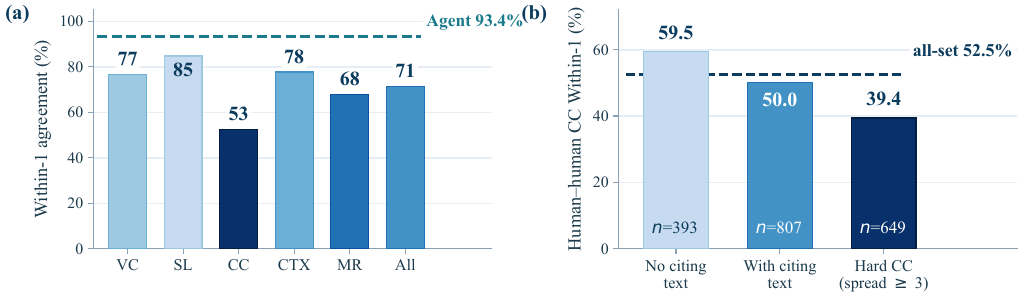}
\caption{(a)~Pairwise human Within-1 by dimension (Table~\ref{tab:human-ceiling}), not the same quantity as model W-1.
Dashed line: best Agent overall W-1 versus \emph{mean} gold in Table~\ref{tab:protocol-ablation} ($93.4\%$).
(b)~Pairwise human CC Within-1 is lower with citing text ($50.0\%$, $n{=}807$) than without ($59.5\%$, $n{=}393$), and $39.4\%$ on items with CC spread $\ge 3$ ($n{=}649$). The with/without split is observational.}
\label{fig:diag}
\end{figure}

\vspace{0.5em}
\noindent Figure~\ref{fig:diag}(a) plots that pairwise ceiling, with the dashed line showing Agent W-1 versus mean gold, a different quantity.
Figure~\ref{fig:diag}(b) tests citing text on gold labels: human CC Within-1 is lower with citing text ($50.0\%$, $n{=}807$) than without ($59.5\%$, $n{=}393$).
The split is observational: figures with citing text may already be harder. $\mathcal{T}$ is not a shortcut for caption agreement, even though index-resolved $\mathcal{T}$ remains necessary to score CTX.

\subsection{Analysis and Discussion}
\label{sec:analysis}

Where a backend is run under all three protocols, overall MAE falls from Direct to Sidecar to Agent, but Agent uses three calls per figure and the Direct-best to Agent-best gap is only $0.025$.
Sidecar supplies OCR and layout cues in one call. Agent separates vision from language and lets the Runner lock VC/SL and cap CC, CTX, and MR.
{\color{blue}The fusion prompt already copies VC and SL and states those caps, so the number that belongs to the Runner is the closed-form MR.} We do not claim the gap is independent of model class or compute.
The drop is uneven across backends. GLM-4.6V falls from Direct MAE $0.662$ to Agent MAE $0.574$, while Qwen-VL-Max still has CC MAE $1.007$ under Agent.
A constant predictor of the \textit{eval1200} mean $8.13$ already reaches MAE $0.565$ and Within-1 $87.9\%$.
Gold scores sit near $8$, so Within-1 barely separates protocols. Agent Within-1 $93.4\%$ is fit to mean gold. Pairwise human overall Within-1 is $71.5\%$, and CC remains the least reliable construct (pairwise Within-1 $52.5\%$, $\alpha{=}0.49$).
{\color{blue}Table~\ref{tab:apx-cc-reason} splits that disagreement into mismatch, omission, and no noted CC defect. Omission is the larger share of the hard figures in that table, so the modest agreement comes from how fully a caption should name the drawn content. The rubric scores an omission at 5 to 6 and a contradiction at 3 to 4.}
These two Within-1 rates ask different questions. $93.4\%$ is how often a model score falls within one point of the rater mean, and $71.5\%$ is how often two raters fall within one point of each other.
Citing text does not close the caption gap. Human CC Within-1 is $50.0\%$ when a citing paragraph is present ($n{=}807$) and $59.5\%$ when it is absent ($n{=}393$). Cited figures may already be harder, so the split is observational. A citing paragraph makes CTX evaluable. It does not supply the panel names, axis titles, or legend entries that CC checks in the caption.
Appendix~\ref{app:article-cases} reads the same split inside five articles, each with every released figure. Overall gold stays near $8.13$, while CC moves when the caption names the panels, axes, or legend entries and when it does not. CTX can stay high on the same figure if the citing paragraph states the claim the caption leaves out.
The open problem is caption and figure verification under manuscript-linked context, not a small MAE protocol delta. Constant-mean floors and backend-specific behavior are in Appendix~\ref{app:extended-analysis}.

\section{Conclusion}

We introduce SciFigQual-Bench for published CS figures and SFQ-Agent, a staged protocol that fuses visual, caption, and citing-text evidence into five-dimensional scores.
On \textit{eval1200}, SFQ-Agent (F3) has the lowest MAE among Direct, Sidecar, and the constant-mean baseline versus mean gold.
Anchoring each figure to its caption and citing paragraphs shifts inspection from isolated visual QA to context-aware reasoning.
Caption verification remains the main gap, and model Within-1 should not replace human raters.

\subsection*{AI use statement}

The experimental protocols query third-party vision and language models as judges (Direct, Sidecar, and SFQ-Agent backends). That usage is the object of study and is reported in Sections~\ref{sec:agent} and~\ref{sec:experiments}, not an undeclared authoring aid.

We used generative AI tools to edit wording for readability.
All technical claims, numerical results, figures, and bibliographic records were checked by the authors, and we take responsibility for the final content of this work.

\subsection*{Ethics statement}

This work studies figures that already appear in publicly available CS conference PDFs (ACL, EMNLP, ICML, NeurIPS, 2020 to 2025) and does not collect private user data or scrape non-public manuscripts.
Trained human annotators scored figure quality under a written rubric, inspecting published figures, captions, and citing excerpts rather than personal or sensitive attributes. Their professional credentials were not collected or disclosed, so we do not describe the ratings as expert judgments. No identifiable human-subject images or private demographic data are released.
Public availability does not by itself grant redistribution rights. The release therefore provides paper identifiers, hashes, figure indices, extraction coordinates, derived annotations, and reconstruction code for all records. Figure crops and text excerpts are redistributed only where the source license permits, otherwise users reconstruct them from the publisher or author-hosted source.
A potential misuse is to treat automated figure scores as a sole accept/reject signal in peer review. SFQ-Agent is intended as an assistive, auditable diagnostic, not a replacement for human reviewers.
Original annotations and code are released under a research-use license, subject to the source-paper licensing constraints above. The project page and code repository are listed in the abstract.

\subsection*{Reproducibility statement}

We take the following steps so that the benchmark and judge protocols can be reproduced.
Section~\ref{sec:task} defines the instance tuple, five-dimensional rubric, and {\color{blue}evidence-availability gating} (Eq.~\ref{eq:overall}). Appendix~\ref{app:rubric} and Appendix~\ref{app:gating} give operational anchors and the evaluability mask.
Section~\ref{sec:construction} and Appendix~\ref{app:curation} specify corpus acquisition, extraction, curation thresholds, and index-driven context binding. Appendix~\ref{app:context} records prompt payload limits.
Appendix~\ref{app:eval1200} defines the frozen \textit{eval1200} split. Appendix~\ref{app:runids} maps run IDs. Table~\ref{tab:protocol-ablation} is the frozen 29-run matrix used for all model MAE/W-1 in the main text. Appendix~\ref{app:measurement} defines the human-ceiling metrics. Table~\ref{tab:human-ceiling} and Figure~\ref{fig:diag} use gold annotations in eval1200/figures.jsonl and do not recompute model scores from that file.
Protocol coverage, constant-mean floors, and split composition are in Appendix~\ref{app:full-leaderboard} and Appendix~\ref{app:measurement}.
Section~\ref{sec:agent} and Appendix~\ref{app:prompts} describe the three protocols. Algorithm~\ref{alg:runner} is the deterministic Runner that locks score ownership and MR fusion.
Appendix~\ref{app:annotation} documents rater workflow, gold aggregation (Eqs.~\ref{eq:dim-mean} and \ref{eq:gold-overall}), and inter-rater dispersion.
Appendix~\ref{app:article-cases} reports five further article-level figure sets with per-rater scores.
{\color{blue}Appendix~\ref{app:faithfulness} checks Direct and Agent explanations as claims on 100 figures from \textit{eval1200}, and prints three of the figures with the claim labels.}
Code and data are released at the project page and GitHub repository listed in the abstract.

\bibliography{scifigqual}
\bibliographystyle{iclr2027_conference}

\clearpage
\appendix
\setcounter{table}{0}
\setcounter{figure}{0}
\setcounter{algorithm}{0}
\renewcommand{\thetable}{A\arabic{table}}
\renewcommand{\thefigure}{A\arabic{figure}}
\renewcommand{\thealgorithm}{A\arabic{algorithm}}
\providecommand{\theHtable}{\thetable}
\providecommand{\theHfigure}{\thefigure}
\renewcommand{\theHtable}{A.\arabic{table}}
\renewcommand{\theHfigure}{A.\arabic{figure}}

\begin{center}
{\Large\bfseries Supplementary Material}\\[0.35em]
{\large Appendix}
\end{center}
\vspace{0.8em}

\section{Benchmark Construction Details}
\label{app:benchmark}

\subsection{Five-Dimensional Rubric}
\label{app:rubric}

In SciFigQual-Bench, all manual scoring rules are uniform: every score is an integer on $[1,10]$, VC, SL, CC, and CTX increase with quality, and MR uses the same range with positive polarity (higher means a careful reader is less likely to be misled).
Dimensions are independent, so a sharp figure can still fail CC/CTX and a weak caption must not automatically lower VC/SL.
Direct, Sidecar, and SFQ-Agent reuse one rubric text and differ only in how evidence is collected. Humans and all models share that rubric, with only the evidence path changing.

All human raters and models follow the same bands: 9 to 10 means near-absence of relevant defects, 7 to 8 minor defects that do not block the main reading, 5 to 6 noticeable friction, 3 to 4 serious impedance, and 1 to 2 near-unusable or severely misleading content.
On CC, CTX, and MR, unresolved ties between adjacent scores are resolved toward the lower score.
The 5 to 8 bands describe mixed-quality cases in the rubric, while empirical scores on this published-only pool concentrate near 8 (Table~\ref{tab:apx-pool-means}).

Table~\ref{tab:apx-rubric-anchors} lists operational anchors.
VC checks publication-scale legibility (blur, contrast, ticks, legends).
SL checks reading path, panel labels, spacing, and chartjunk~\citep{tufte2001}.
CC checks whether caption $c$ names the right objects, metrics, panels, and trends.
CTX checks whether citing paragraphs $\mathcal{T}$ make claims that $I$ supports, and is not a writing-quality score.
MR checks truncated axes, missing baselines/error bars, unfair comparison, and visual and text conflicts~\citep{vlmviz2025}, kept separate from CC/CTX so that a vague caption alone does not collapse trust when the plot is fair.

\begin{table}[ht]
\caption{Operational anchors for the five scoring dimensions.}
\label{tab:apx-rubric-anchors}
\begin{center}
\small
\setlength{\tabcolsep}{3pt}
\renewcommand{\arraystretch}{1.05}
\begin{tabular}{@{}p{0.08\linewidth}p{0.28\linewidth}p{0.28\linewidth}p{0.28\linewidth}@{}}
\toprule
\rowcolor{tabheader}
\textbf{Dim.} & \textbf{Primary cues} & \textbf{High band} & \textbf{Low band} \\
\midrule
VC & Resolution, blur, axis/legend legibility & Readable at publication scale & Effortful (5 to 6). Illegible key labels (3 to 4). Unreadable (1 to 2) \\
\rowcolor{tabalt}
SL & Whitespace, panel order, $(a)$/$(b)$, flow & Clear path and complete labels & Costly structure (5 to 6). Disordered (3 to 4). No hierarchy (1 to 2) \\
CC & Entities, metrics, panels, numeric alignment & Covers objects/metrics/panels & Vague (5 to 6). Mismatch (3 to 4). Wrong-figure (1 to 2) \\
\rowcolor{tabalt}
CTX & Index, trends/comparisons vs.\ $I$ & Specific evidentiary match & Bare pointer (5 to 6). Wrong claim/figure (3 to 4 / 1 to 2) \\
MR & Axes, baselines, fairness, conflicts & Trustworthy and fair & Ambiguous (5 to 6). Missing signals (3 to 4). Contradiction (1 to 2) \\
\bottomrule
\end{tabular}
\end{center}
\end{table}

To prevent fluent but content-empty text from receiving inflated scores, we impose hard caps on CC and CTX.
A caption that describes the figure but omits the key metric or compared groups receives CC at most 5 to 6 (cap~$=6$), and a citation that merely points to the figure without articulating any finding (e.g., ``see Figure~3'') is capped at 5 to 6 on CTX, never above 6.
By contrast, a citation that reverses the actual trend receives 1 to 2 on both CTX and MR.
These caps apply independently of visual quality: a visually pristine figure cannot compensate for substantive textual misalignment, OCR- or CV-derived evidence may support VC/SL but cannot override the pixel-level content, and inapplicable sub-criteria are ignored.
{A cap is a ceiling, not a required score.
Each rater applies it only after judging whether the key metric or the compared groups are missing for that figure.
Raters who differ on that judgment return different scores and still follow the same rubric.
Released gold is the dimension-first mean of the independent ratings (Eqs.~\ref{eq:dim-mean} and \ref{eq:gold-overall}).
The mean is not clipped, so a score from a rater who judges the metric as already specified stays in the gold.
Case~4, Figure~3 in Appendix~\ref{app:article-cases} is a recorded instance.
}

\subsection{\texorpdfstring{{Evidence-Availability Gating}}{Evidence-Availability Gating}}
\label{app:gating}

Missing metadata must not be scored as low quality.
Table~\ref{tab:apx-gating} is the mask shared by human gold and every model protocol: if both $c$ and $\mathcal{T}$ are empty the instance is dropped. Otherwise VC/SL/MR stay available from the image, CC requires a caption, CTX requires citing text, and the overall is the arithmetic mean of the remaining dimension scores.

\begin{table}[ht]
\caption{{Evidence-availability gating} of evaluable dimensions.}
\label{tab:apx-gating}
\begin{center}
\small
\setlength{\tabcolsep}{6pt}
\begin{tabular}{@{}lccccc@{}}
\toprule
\rowcolor{tabheader}
\textbf{Evidence} & \textbf{VC} & \textbf{SL} & \textbf{CC} & \textbf{CTX} & \textbf{MR} \\
\midrule
$c$ and $\mathcal{T}$ present & yes & yes & yes & yes & yes \\
\rowcolor{tabalt}
only $c$ & yes & yes & yes & null & yes \\
only $\mathcal{T}$ & yes & yes & null & yes & yes \\
\rowcolor{tabalt}
both absent & \multicolumn{5}{c}{instance excluded} \\
\bottomrule
\end{tabular}
\end{center}
\end{table}

\subsection{Curation Stages S1 and S5}
\label{app:curation}

Raw layout extraction yields many non-evaluable crops.
We therefore run five sequential stages that separate hard deletion from soft retention (Table~\ref{tab:apx-curation}).
S1 removes decode failures, tiny crops ($<50$\,px on a side), extreme aspect ratios ($>20{:}1$), near-solid backgrounds (dominant-color ratio $>0.97$), and near-duplicate pHashes (Hamming $\le 8$).
S2 only \emph{flags} anomalies within a paper (area $>3\times$ the paper median, height or width overflow, or an abnormal figure index), without deleting rows.
S3 re-crops from the PDF with caption-anchored boxes when flags indicate broken extraction. Failed recoveries fall back to the original crop and are marked rather than silently dropped.
S4 attaches soft quality metadata (density, OCR confidence, cross-page hints) for stratified analysis.
S5 drops papers whose unusable-figure rates exceed fixed thresholds (e.g., hard-delete ratio $>0.30$), so context binding is not attempted on irrecoverable layouts.
A figure must survive S1 and S3 to enter the released set. S4 keeps borderline cases. S5 acts at paper granularity.
The funnel from $K_0{=}62{,}694$ PDFs to $P{=}1{,}144$ papers and $N{=}7{,}609$ figures is the result of these filters, not random subsampling.

\begin{table}[ht]
\caption{Curation stages and primary actions (implementation thresholds).}
\label{tab:apx-curation}
\begin{center}
\small
\setlength{\tabcolsep}{4pt}
\begin{tabular}{@{}cp{0.22\linewidth}p{0.58\linewidth}@{}}
\toprule
\rowcolor{tabheader}
\textbf{S} & \textbf{Name} & \textbf{Primary action} \\
\midrule
1 & Hard filter & Delete decode failures, $<50$\,px, aspect $>20$, solid $>0.97$, pHash dup.\ $\le8$ \\
\rowcolor{tabalt}
2 & Anomaly flag & Soft-flag oversized / overflow / bad-index crops (no delete) \\
3 & Re-extract & Caption-anchored PDF re-crop for flagged failures. Else keep original \\
\rowcolor{tabalt}
4 & Soft mark & Attach quality metadata for later stratification \\
5 & Paper filter & Drop papers with repeated unusable figures / missing indices \\
\bottomrule
\end{tabular}
\end{center}
\end{table}

\subsection{Context Binding and Prompt Payloads}
\label{app:context}

Citing bundles $\mathcal{T}_k$ are resolved from figure-index patterns in body text (e.g., ``Figure~$N$'' / ``Fig.~$N$''), not from abstract heuristics.
Matched paragraphs are merged within section boundaries. Blocks shorter than 20 characters are ignored.
For scoring prompts we further truncate payloads so that backends see comparable context: up to two snippets of at most 500 characters for the target figure, plus an optional paper-level figure map with shortened captions ($\le220$ characters).
After aggregation, bare-pointer paragraphs are down-weighted and comparative/trend language is preferred for CTX, but we never invent citations. Absent index matches leave CTX gated null.
Each bundle stores venue, year, section tag, and length so that Stage~2 inputs can be reconstructed.

{\subsection{Corpus Acquisition and Layout Parsing}
\label{app:acquisition}

PDFs and bibliographic metadata are crawled from OpenReview, ACL Anthology, and PMLR, and each file is tagged with $(\mathit{venue},\mathit{year},\mathit{paper\_id})$.
PDFs are linearized, publisher watermarks are stripped where possible, and exact duplicates are removed by the content fingerprint $h(p)=\mathrm{SHA256}(\mathrm{bytes}(p))$.
Records outside 2020 to 2025 or missing parseable metadata are discarded.
The surviving set is the raw corpus $\mathcal{C}_0$ in Eq.~\ref{eq:corpus}.

Each PDF is rasterized at 300\,DPI and its layout is parsed with Marker, which returns text blocks, figure bounding boxes, and caption candidates.
Candidates are ranked by spatial overlap, vertical distance, and a figure-caption cue~\citep{figex2025}.
The five curation stages that follow are given in Appendix~\ref{app:curation}.
}

\section{Experimental Protocol Details}
\label{app:experiments}

\subsection{eval1200 Split Definition}
\label{app:eval1200}

\textit{eval1200} is the frozen evaluation set for all protocol and backend ablations ($n{=}1{,}200$ figures from the $6{,}308$ rated instances), and each figure has at least three independent human ratings (Table~\ref{tab:apx-rater-n}). Gold labels and protocol contracts were frozen before any model call. The split is released publicly for reproducibility after that freeze. It is not presented as a permanently hidden leaderboard test.
Sampling is paper-aware (figures from the same paper are kept or dropped together), yielding 1,200 figures from 254 papers mixed by figure type rather than forced to equal venue counts. Table~\ref{tab:apx-eval-comp} records venue and year counts and Table~\ref{tab:apx-eval-domain} records topic labels on the same split.
Gold labels are frozen before any model call.
Every Direct / Sidecar / SFQ-Agent configuration consumes identical $(I,c,\mathcal{T},m)$ under the {evidence-availability mask} in Table~\ref{tab:apx-gating}.
The $29$ rows are not a complete $11{\times}3$ factorial: Sidecar omits Claude backends. Agent adds Gemini-3.1-Pro and omits Doubao, Pixtral, and InternVL (Table~\ref{tab:apx-coverage}).

\begin{table}[ht]
\caption{\textit{eval1200} venue and year composition ($n{=}1{,}200$ figures, 254 papers). Venue counts match the main-text split. Years are from the gold JSONL.}
\label{tab:apx-eval-comp}
\begin{center}
\small
\begin{tabular}{@{}lc@{}}
\toprule
\rowcolor{tabheader}
\textbf{Venue} & \textbf{Figures} \\
\midrule
ACL & 266 \\
\rowcolor{tabalt}
EMNLP & 294 \\
ICML & 294 \\
\rowcolor{tabalt}
NeurIPS & 346 \\
\midrule
All & 1{,}200 \\
\bottomrule
\end{tabular}
\hspace{2.4em}
\begin{tabular}{@{}lc@{}}
\toprule
\rowcolor{tabheader}
\textbf{Year} & \textbf{Figures} \\
\midrule
2020 & 155 \\
\rowcolor{tabalt}
2021 & 140 \\
2022 & 197 \\
\rowcolor{tabalt}
2023 & 190 \\
2024 & 272 \\
\rowcolor{tabalt}
2025 & 246 \\
\midrule
All & 1{,}200 \\
\bottomrule
\end{tabular}
\end{center}
\end{table}

\begin{table}[ht]
\caption{Topic labels on \textit{eval1200} ($n{=}1{,}200$).
``Other labeled'' aggregates reinforcement learning, multimodal, graph learning, and speech.
Unlabeled items have no topic tag in the release.}
\label{tab:apx-eval-domain}
\begin{center}
\small
\setlength{\tabcolsep}{8pt}
\begin{tabular}{@{}lcc@{}}
\toprule
\rowcolor{tabheader}
\textbf{Topic} & \textbf{Figures} & \textbf{Share} \\
\midrule
Natural language processing & 535 & 44.6\% \\
\rowcolor{tabalt}
Machine learning (general) & 360 & 30.0\% \\
Computer vision & 139 & 11.6\% \\
\rowcolor{tabalt}
Other labeled CS topics & 77 & 6.4\% \\
Unlabeled & 89 & 7.4\% \\
\midrule
All & 1{,}200 & 100\% \\
\bottomrule
\end{tabular}
\end{center}
\end{table}

\subsection{Run ID Mapping}
\label{app:runids}

Each configuration has a stable run ID and call budget.
Direct (1 call/figure): D1 Gemini-3.5-Flash, D2 GPT-5.6-Sol, D3 Claude-Sonnet-5, D4 Qwen-VL-Max, D5 GLM-4.6V, D6 Doubao-Seed-2.0-pro, D7 Llama-4-Maverick, D8 Pixtral-Large, D9 Nova-Pro, D10 Claude-Opus-4.8, D11 InternVL3-78B.
Sidecar (1 call/figure): S1 Gemini-3.5-Flash, S2 GPT-5.6-Sol, S3 Qwen-VL-Max, S4 GLM-4.6V, S5 Doubao-Seed-2.0-pro, S6 Llama-4-Maverick, S7 Pixtral-Large, S8 Nova-Pro, S9 InternVL3-78B, each with PaddleOCR-VL~\citep{paddleocrvl2025} side features.
SFQ-Agent (3 calls/figure): F1 Gemini-3.5-Flash/Gemini-3.5-Flash, F2 Gemini-3.1-Pro/Gemini-3.1-Pro, F3 GPT-5.6-Sol/GPT-5.6-Sol, F4 Claude-Sonnet-5/Claude-Sonnet-5, F5 Qwen-VL-Max+Qwen-Plus, F6 GLM-4.6V/GLM-4.6V, F7 Llama-4-Maverick/Llama-4-Maverick, F8 Claude-Opus-4.8/Claude-Opus-4.8, F9 Nova-Pro/Nova-Pro.
Slash notation denotes matched vision/language backends. F5 is the only heterogeneous pairing.
Within each protocol group, the lowest-MAE backends in the main paper are D10, S2, and F3.

\subsection{Protocol and Backend Coverage}
\label{app:full-leaderboard}

The 29-run MAE/W-1 matrix is Table~\ref{tab:protocol-ablation} in the main text (the frozen copy used for all model scores there).
Table~\ref{tab:apx-coverage} records which backends were run under each protocol. Empty cells are not imputed.
Direct and Sidecar use one call per figure. Agent uses three.

\begin{table}[ht]
\caption{Run-ID coverage of the 29 configurations.
Empty cells were not run.
F5 is the only heterogeneous vision/language pairing (Qwen-VL-Max + Qwen-Plus).}
\label{tab:apx-coverage}
\begin{center}
\small
\setlength{\tabcolsep}{6pt}
\begin{tabular}{@{}lccc@{}}
\toprule
\rowcolor{tabheader}
\textbf{Backend} & \textbf{Direct} & \textbf{Sidecar} & \textbf{Agent} \\
\midrule
Gemini-3.5-Flash & D1 & S1 & F1 \\
\rowcolor{tabalt}
Gemini-3.1-Pro & n/a & n/a & F2 \\
GPT-5.6-Sol & D2 & S2 & F3 \\
\rowcolor{tabalt}
Claude-Sonnet-5 & D3 & n/a & F4 \\
Qwen-VL-Max & D4 & S3 & F5 \\
\rowcolor{tabalt}
GLM-4.6V & D5 & S4 & F6 \\
Doubao-Seed-2.0-pro & D6 & S5 & n/a \\
\rowcolor{tabalt}
Llama-4-Maverick & D7 & S6 & F7 \\
Pixtral-Large & D8 & S7 & n/a \\
\rowcolor{tabalt}
Nova-Pro & D9 & S8 & F9 \\
Claude-Opus-4.8 & D10 & n/a & F8 \\
\rowcolor{tabalt}
InternVL3-78B & D11 & S9 & n/a \\
\bottomrule
\end{tabular}
\end{center}
\end{table}

\subsection{Human-Ceiling and Slice Metrics}
\label{app:measurement}

Table~\ref{tab:human-ceiling} and Figure~\ref{fig:diag} use gold annotations in eval1200/figures.jsonl only. They do not restate model MAE/W-1 from Table~\ref{tab:protocol-ablation}.
Table~\ref{tab:apx-metric-defs} distinguishes pairwise human agreement from model agreement versus mean gold.
Human leave-one-out MAE compares each rater to the mean of the other raters on that figure and dimension. Pairwise Within-1 averages all rater pairs.
Ordinal Krippendorff's $\alpha$ uses coincidence counts on integer scores in $\{1,\ldots,10\}$.
Hard CC is the subset whose per-figure CC spread is at least 3 points ($n{=}649$).
Of $807$ figures with citing text, $804$ retain $\ge$2 CTX ratings after per-rater nulls (three keep only one; Table~\ref{tab:human-ceiling}). Figure~\ref{fig:diag}(b) reports the $807$/$393$ citing split, restated numerically in Table~\ref{tab:apx-cc-slices}.
The dashed line in Figure~\ref{fig:diag}(a) is the best Agent overall W-1 from Table~\ref{tab:protocol-ablation} ($93.4\%$), which is versus \emph{mean} gold rather than pairwise human agreement.
A constant predictor of each dimension's \textit{eval1200} mean is reported in Table~\ref{tab:apx-constant}. Overall MAE is $0.565$ and W-1 is $87.9\%$ versus mean $8.13$.
Table~\ref{tab:apx-overall-floor} places that floor next to the best row of each protocol in Table~\ref{tab:protocol-ablation}.

\begin{table}[ht]
\caption{Metric definitions used in the main text.
Model scores always follow Table~\ref{tab:protocol-ablation}. Human-ceiling numbers follow gold JSONL only.}
\label{tab:apx-metric-defs}
\begin{center}
\small
\setlength{\tabcolsep}{4pt}
\begin{tabular}{@{}p{0.22\linewidth}p{0.48\linewidth}p{0.22\linewidth}@{}}
\toprule
\rowcolor{tabheader}
\textbf{Quantity} & \textbf{Definition} & \textbf{Where used} \\
\midrule
Pairwise W-1 & Fraction of rater pairs with $|s_r-s_{r'}|\le 1$ & Table~\ref{tab:human-ceiling}. Figure~\ref{fig:diag} \\
\rowcolor{tabalt}
Leave-one-out MAE & $|s_r-\mathrm{mean}_{r'\neq r}s_{r'}|$ averaged over raters and items & Table~\ref{tab:human-ceiling} \\
Ordinal $\alpha$ & Krippendorff $\alpha$ on integer scores $\{1,\ldots,10\}$. Items with $<2$ ratings dropped & Table~\ref{tab:human-ceiling} \\
\rowcolor{tabalt}
Model MAE / W-1 / SRCC & Versus dimension-first \emph{mean} gold (Eqs.~\ref{eq:dim-mean} and \ref{eq:gold-overall}) & Table~\ref{tab:protocol-ablation} \\
Constant MAE / W-1 & Predict the \textit{eval1200} mean of that dimension & Table~\ref{tab:apx-constant} \\
\bottomrule
\end{tabular}
\end{center}
\end{table}

\begin{table}[ht]
\caption{Constant-mean predictor on \textit{eval1200} gold (no model outputs).
Each row predicts that dimension's empirical mean.
CTX is restricted to the $807$ figures with citing text.
SRCC is undefined for a constant.}
\label{tab:apx-constant}
\begin{center}
\small
\setlength{\tabcolsep}{6pt}
\begin{tabular}{@{}lcccc@{}}
\toprule
\rowcolor{tabheader}
\textbf{Dim.} & \textbf{Mean} & \textbf{$n$} & \textbf{MAE}$\downarrow$ & \textbf{W-1}$\uparrow$ \\
\midrule
VC & 8.31 & 1{,}200 & 0.572 & 83.3\% \\
\rowcolor{tabalt}
SL & 8.61 & 1{,}200 & 0.404 & 95.4\% \\
CC & 7.48 & 1{,}200 & 1.092 & 53.2\% \\
\rowcolor{tabalt}
CTX & 8.27 & 807 & 0.989 & 60.2\% \\
MR & 8.04 & 1{,}200 & 0.778 & 79.4\% \\
\midrule
Overall & 8.13 & 1{,}200 & 0.565 & 87.9\% \\
\bottomrule
\end{tabular}
\end{center}
\end{table}

\begin{table}[ht]
\caption{Overall \textit{eval1200} constant floor versus the best row of each protocol in Table~\ref{tab:protocol-ablation}.
Constant and model W-1 are both versus \emph{mean} gold, not pairwise human W-1.}
\label{tab:apx-overall-floor}
\begin{center}
\small
\setlength{\tabcolsep}{6pt}
\begin{tabular}{@{}lccc@{}}
\toprule
\rowcolor{tabheader}
\textbf{Predictor} & \textbf{MAE}$\downarrow$ & \textbf{W-1}$\uparrow$ & \textbf{SRCC}$\uparrow$ \\
\midrule
Constant mean ($8.13$) & 0.565 & 87.9\% & n/a \\
\rowcolor{tabalt}
Direct-best (D10) & 0.443 & 92.4\% & 0.582 \\
Sidecar-best (S2) & 0.436 & 92.6\% & 0.578 \\
\rowcolor{tabbest}
Agent-best (F3) & 0.418 & 93.4\% & 0.598 \\
\bottomrule
\end{tabular}
\end{center}
\end{table}

\begin{table}[ht]
\caption{Pairwise human CC Within-1 on \textit{eval1200} slices (gold only). Companion to Figure~\ref{fig:diag}(b).
The with/without citing-text split is observational.}
\label{tab:apx-cc-slices}
\begin{center}
\small
\setlength{\tabcolsep}{8pt}
\begin{tabular}{@{}lcc@{}}
\toprule
\rowcolor{tabheader}
\textbf{Slice} & \textbf{$n$} & \textbf{Pairwise CC W-1} \\
\midrule
All \textit{eval1200} & 1{,}200 & 52.5\% \\
\rowcolor{tabalt}
No citing text & 393 & 59.5\% \\
With citing text & 807 & 50.0\% \\
\rowcolor{tabalt}
Hard CC (spread $\ge 3$) & 649 & 39.4\% \\
\bottomrule
\end{tabular}
\end{center}
\end{table}

{A keyword pass over the released per-rater CC rationales splits the same $1{,}200$ figures by what the raters wrote (Table~\ref{tab:apx-cc-reason}).
Explicit mismatch, omission only, and no noted CC defect have pairwise CC W-1 $41.7\%$, $46.3\%$, and $67.7\%$, and mean spread $3.59$, $3.01$, and $1.87$.
Among the $649$ figures with a CC spread of at least $3$, $435$ are omission-only and $96$ contain a mismatch claim.
Case~4, Figure~3 in Appendix~\ref{app:article-cases} is an omission cap: one rater scores CC $6$, and gold CC $8.33$ is the unclipped mean of $10$, $9$, and $6$.
Ex-H in Appendix~\ref{app:model-cases} is a contradiction: the caption names a red bar the figure does not draw, and gold CC is $3$.

\begin{table}[ht]
\caption{CC rationales on \textit{eval1200}, grouped by a keyword pass. The three groups partition the $1{,}200$ figures.}
\label{tab:apx-cc-reason}
\begin{center}
\small
\setlength{\tabcolsep}{6pt}
\begin{tabular}{@{}lccc@{}}
\toprule
\rowcolor{tabheader}
\textbf{Group} & \textbf{$n$} & \textbf{Pairwise CC W-1} & \textbf{Mean spread} \\
\midrule
Explicit mismatch & 138 & 41.7\% & 3.59 \\
\rowcolor{tabalt}
Omission only & 613 & 46.3\% & 3.01 \\
No noted CC defect & 449 & 67.7\% & 1.87 \\
\bottomrule
\end{tabular}
\end{center}
\end{table}

\begin{table}[ht]
\caption{Model CC MAE on the Table~\ref{tab:apx-cc-reason} omission-only and mismatch groups for the main Direct (D2) and Agent (F3) GPT-5.6-Sol runs.
The table covers the $751$ figures with a human-disagreement label.
Within this split, omission-only accounts for $82\%$ (D2) and $81\%$ (F3) of the aggregate absolute CC error in this subset.
Over all $1{,}200$ figures the corresponding shares are $62\%$ and $60\%$.
The remaining $449$ figures without a human-disagreement label have lower MAE (D2 $0.618$, F3 $0.615$), so the full-pool means in Table~\ref{tab:protocol-ablation} sit below either disagreement-group MAE.}
\label{tab:apx-cc-model-mae}
\begin{center}
\small
\setlength{\tabcolsep}{6pt}
\begin{tabular}{@{}lccc@{}}
\toprule
\rowcolor{tabheader}
\textbf{Group} & \textbf{$n$} & \textbf{D2 CC MAE} & \textbf{F3 CC MAE} \\
\midrule
Omission only & 613 & 1.110 & 1.051 \\
\rowcolor{tabalt}
Explicit mismatch & 138 & 1.076 & 1.087 \\
\bottomrule
\end{tabular}
\end{center}
\end{table}
}

{\begin{table}[ht]
\caption{Same-backbone ReAct versus SFQ-Agent on 120 figures with local images.
The subset has 30 figures from each of ACL, EMNLP, ICML, and NeurIPS.
Compute and time limits in the rebuttal window restrict this release to the subset above. The same comparison will be extended to the full 1,200 figures in a follow-up.
Doubao-Seed-2.1-Pro uses \texttt{doubao-seed-2-1-pro-260628}. GPT-5.6-Sol uses \texttt{gpt-5.6-sol}.
Temperature is 0 and extended reasoning is disabled for every call.
Both backbones receive the same official prompts.
ReAct follows the paper Direct scoring prompt and does not use the Runner.
SFQ-Agent uses the paper's three stage prompts, locks VC and SL from the vision call, and fuses MR with the Runner.
Neither prompt is rewritten or shortened for this comparison.
Neither protocol receives OCR or CV features.
Both protocols completed all 120 figures.
API usage on 40 of the 120 figures, under the same prompts and three-call schedule, gives Doubao-Seed-2.1-Pro mean calls of 3.73 for ReAct and 3.00 for SFQ-Agent, with mean tokens 11363 and 10913.
GPT mean calls are 3.40 and 3.00, with mean tokens 6412 and 8208.
Context consistency uses the 54 figures on each backbone that have a human context score.
MAE is against the human scores. Lower is better.
This comparison stays outside Table~\ref{tab:protocol-ablation}.
Its MAE does not replace the \textit{eval1200} MAE in that table.
Doubao-Seed-2.1-Pro is separate from Doubao-Seed-2.0-pro.}
\label{tab:apx-react-sfq}
\begin{center}
\small
\setlength{\tabcolsep}{4pt}
\begin{tabular}{@{}lcccc@{}}
\toprule
\rowcolor{tabheader}
\textbf{Dimension} & \textbf{Seed-2.1 ReAct} & \textbf{Seed-2.1 SFQ} & \textbf{GPT ReAct} & \textbf{GPT SFQ} \\
\midrule
Overall & 1.461 & 0.873 & 1.493 & 0.746 \\
\rowcolor{tabalt}
Visual clarity & 0.962 & 0.576 & 0.928 & 0.578 \\
Structure layout & 0.840 & 0.568 & 0.865 & 0.582 \\
\rowcolor{tabalt}
Caption consistency & 1.866 & 1.199 & 1.556 & 1.124 \\
Context consistency & 2.343 & 1.176 & 4.731 & 0.719 \\
\rowcolor{tabalt}
Misleading risk & 2.001 & 1.601 & 1.674 & 1.593 \\
\bottomrule
\end{tabular}
\end{center}
\end{table}
}

\section{Judge Prompts and Deterministic Runner}
\label{app:prompts}

Direct and Sidecar share a single-pass template. Sidecar additionally receives OCR/CV features.
SFQ-Agent issues three staged calls, then applies the code-level Runner in Algorithm~\ref{alg:runner}.
Full prompt templates are included in the public code release (\url{https://github.com/FrankDengAI/SciFigQual-Bench}).

\subsection{Single-Pass Judges (Direct and Sidecar)}
\label{app:direct-prompt}

Direct: one VLM call over $(I,c,\mathcal{T})$ returns five integer scores with short evidence-tied reasons. Gated-off CC/CTX are null.
Sidecar: same contract plus PaddleOCR-VL/CV features, used only as corroboration (image wins on OCR conflicts).
Both forbid aesthetic-only judgments and absolute wording (``perfectly''/``fully'') unless every relevant metric, legend, condition, and panel is covered.

\subsection{SFQ-Agent Evidence Stages}
\label{app:agent-prompts}

\paragraph{Vision.}
VLM reads $I$+features, \emph{writes final} VC/SL, and emits image facts plus visual risk band $[\ell,h]$. It must not score CC/CTX/final MR.

\paragraph{Language.}
LLM reads only $c$/$\mathcal{T}$ (no pixels) and emits \emph{text-side} proposals: omissions/bare-reference caps $\mathrm{cap}_{\mathrm{CC}},\mathrm{cap}_{\mathrm{CTX}}$ (typically $6$) and a {text-claim severity}
$\sigma\in\{\mathrm{none},\mathrm{minor},\mathrm{moderate},\mathrm{severe},\mathrm{contradiction}\}$
as a prior over caption/context friction, not a pixel-verified cross-modal verdict. No final scores.

\paragraph{Fusion.}
Judge proposes CC/CTX and an MR \emph{rationale} from the two evidence reports (any numeric MR it emits is only a draft).
The Runner then \emph{overwrites} VC/SL from Stage~1 and \emph{replaces} the MR number by Algorithm~\ref{alg:runner}'s closed form of $[\ell,h]$ and $\sigma$, while applying $\mathrm{cap}_{\mathrm{CC}/\mathrm{CTX}}$ and an optional MR cap from Stage~2, which is usually null and is set only for an explicit severe contradiction that must not receive a high MR.

\subsection{Runner Hard Constraints}
\label{app:runner}

{Section~\ref{sec:agent} gives the reason for separating the two calls, the score each call may write, and the deterministic Runner in Algorithm~\ref{alg:runner}.
The paragraphs above state the prompt contract, and the supplement contains the full templates.
}

\section{Human Annotation Protocol}
\label{app:annotation}

\subsection{Rater Workflow}
\label{app:rater-workflow}

Annotators inspect $I$ first, then $c$, then $\mathcal{T}$, and only afterwards optional OCR/CV features. Model predictions are never shown, so humans are not anchored by automated scores.
Each dimension requires a short rationale grounded in visible marks or exact phrases, and plausible but unfaithful explanations are rejected~\citep{faithfulnlp2020}.
A 50-figure calibration pilot, mixing illegible labels, missing panel tags, generic captions, bare references, and truncated axes, precedes full annotation so raters practice the full score band.

\subsection{Quality Control and Dimension Statistics}
\label{app:qc}

Dual annotation on an ACL 2025 holdout ($n{=}281$ figures, 48 papers) checked rubric operability before scaling, and overall disagreements exceeding two points on that holdout triggered senior adjudication.
\textit{eval1200} is distinct from that dual holdout: every figure has at least three ratings, 119 figures have additional ratings (Table~\ref{tab:apx-rater-n}), and released gold uses all available ratings under Eqs.~\ref{eq:dim-mean} and \ref{eq:gold-overall}.
Table~\ref{tab:apx-pool-means} compares dimension means on the full rated pool and on \textit{eval1200}. CC is the weakest axis on both, because many published figures are visually adequate yet captions omit metrics or comparison groups.

\begin{table}[ht]
\caption{Number of human ratings per \textit{eval1200} figure.
Gold means and IRR use every available rating on that figure.}
\label{tab:apx-rater-n}
\begin{center}
\small
\setlength{\tabcolsep}{8pt}
\begin{tabular}{@{}lc@{}}
\toprule
\rowcolor{tabheader}
\textbf{\# raters} & \textbf{Figures} \\
\midrule
3 & 1{,}081 \\
\rowcolor{tabalt}
4 & 64 \\
6 & 1 \\
\rowcolor{tabalt}
7 & 1 \\
8 & 50 \\
\rowcolor{tabalt}
9 & 3 \\
\midrule
All & 1{,}200 \\
\bottomrule
\end{tabular}
\end{center}
\end{table}

\begin{table}[ht]
\caption{Mean human scores on the full rated pool versus \textit{eval1200}.
CTX means are over items with citing text ({evidence-available}).}
\label{tab:apx-pool-means}
\begin{center}
\small
\setlength{\tabcolsep}{6pt}
\begin{tabular}{@{}lcc@{}}
\toprule
\rowcolor{tabheader}
\textbf{Dim.} & \textbf{Rated pool} ($n{=}6{,}308$) & \textbf{\textit{eval1200}} ($n{=}1{,}200$) \\
\midrule
VC & 8.12 & 8.31 \\
\rowcolor{tabalt}
SL & 8.58 & 8.61 \\
CC & 7.42 & 7.48 \\
\rowcolor{tabalt}
CTX & 7.89 & 8.27 \\
MR & 8.01 & 8.04 \\
\midrule
Overall & 8.05 & 8.13 \\
\bottomrule
\end{tabular}
\end{center}
\end{table}

\subsection{Gold Standard Aggregation}
\label{app:aggregation}

Released gold is aggregated \emph{dimension-first}, matching the {evidence-availability mask} at inference.
For each evaluable dimension $d$ and figure $i$,
\begin{equation}
\bar{s}_i(d)=\frac{1}{|\{r: s_{i,r}(d)\neq\mathrm{null}\}|}\sum_{r} s_{i,r}(d),
\label{eq:dim-mean}
\end{equation}
then
\begin{equation}
\bar{y}_i=\frac{1}{\sum_{d} g_i(d)}\sum_{d} g_i(d)\,\bar{s}_i(d).
\label{eq:gold-overall}
\end{equation}
On a complete rating matrix (every rater scores the same evaluable dimensions), $\bar{y}_i$ equals the mean of per-rater overalls by linearity.
The two aggregations diverge only when raters have \emph{uneven} {evidence-availability null} masks on the same figure (e.g., one rater nulls CTX while another scores it). The release always uses Eqs.~\ref{eq:dim-mean} and \ref{eq:gold-overall}.

\paragraph{Worked example (Ex-B in Figure~\ref{fig:apx-human-cases}).}
Ex-B receives rater scores $(6,5,5,6,4)$, $(5,6,5,6,4)$, and $(8,9,6,7,5)$.
Per-rater overalls are $5.20$, $5.20$, and $7.00$.
Dimension means are $\bar{s}=(6.33,6.67,5.33,6.33,4.33)$, so $\bar{y}=5.80$, which coincides with the mean of the three per-rater overalls on this complete case. The release uses $5.80$.

\paragraph{Null handling.}
Absent $\mathcal{T}$ nulls CTX for every rater (Ex-A). Absent $c$ nulls CC.
Table~\ref{tab:apx-irr-stats} reports dispersion on \textit{eval1200}: CC has the largest mean spread ($2.65$) and lowest W-1 pair rate ($52.5\%$), aligning with its role as the hardest axis for both humans and models.

\begin{table}[ht]
\caption{Inter-rater dispersion on \textit{eval1200} ($n{=}1{,}200$ figures; at least three raters on VC/SL/CC/MR/overall).
Spread $=\max_r s_{i,r}(d)-\min_r s_{i,r}(d)$ per figure. Median is the median of that spread. W-1 pair rate $=$ fraction of rater pairs within $\pm1$ point.}
\label{tab:apx-irr-stats}
\begin{center}
\small
\setlength{\tabcolsep}{5pt}
\renewcommand{\arraystretch}{1.05}
\begin{tabular}{@{}lcccc@{}}
\toprule
\rowcolor{tabheader}
\textbf{Dimension} & \textbf{Mean spread} & \textbf{Median spread} & \textbf{W-1 pair} & \textbf{$n$} \\
\midrule
VC & 1.52 & 1.0 & 76.6\% & 1{,}200 \\
\rowcolor{tabalt}
SL & 1.03 & 1.0 & 84.8\% & 1{,}200 \\
CC & 2.65 & 3.0 & 52.5\% & 1{,}200 \\
\rowcolor{tabalt}
CTX & 1.54 & 1.0 & 77.7\% & 804 \\
MR & 1.92 & 2.0 & 67.9\% & 1{,}200 \\
\rowcolor{tabalt}
Overall (per-rater) & 1.26 & 1.0 & 71.5\% & 1{,}200 \\
\midrule
\multicolumn{5}{@{}p{0.95\linewidth}@{}}{\textit{eval1200 means:} VC 8.31. SL 8.61. CC 7.48. CTX 8.27. MR 8.04. Overall 8.13. CTX $n{=}804$ items with $\ge$2 ratings ($807$ figures have citing text).} \\
\bottomrule
\end{tabular}
\end{center}
\end{table}

\subsection{Representative Multi-Rater Cases}
\label{app:human-cases}

\begin{figure}[ht]
\begin{center}
\includegraphics[width=\linewidth]{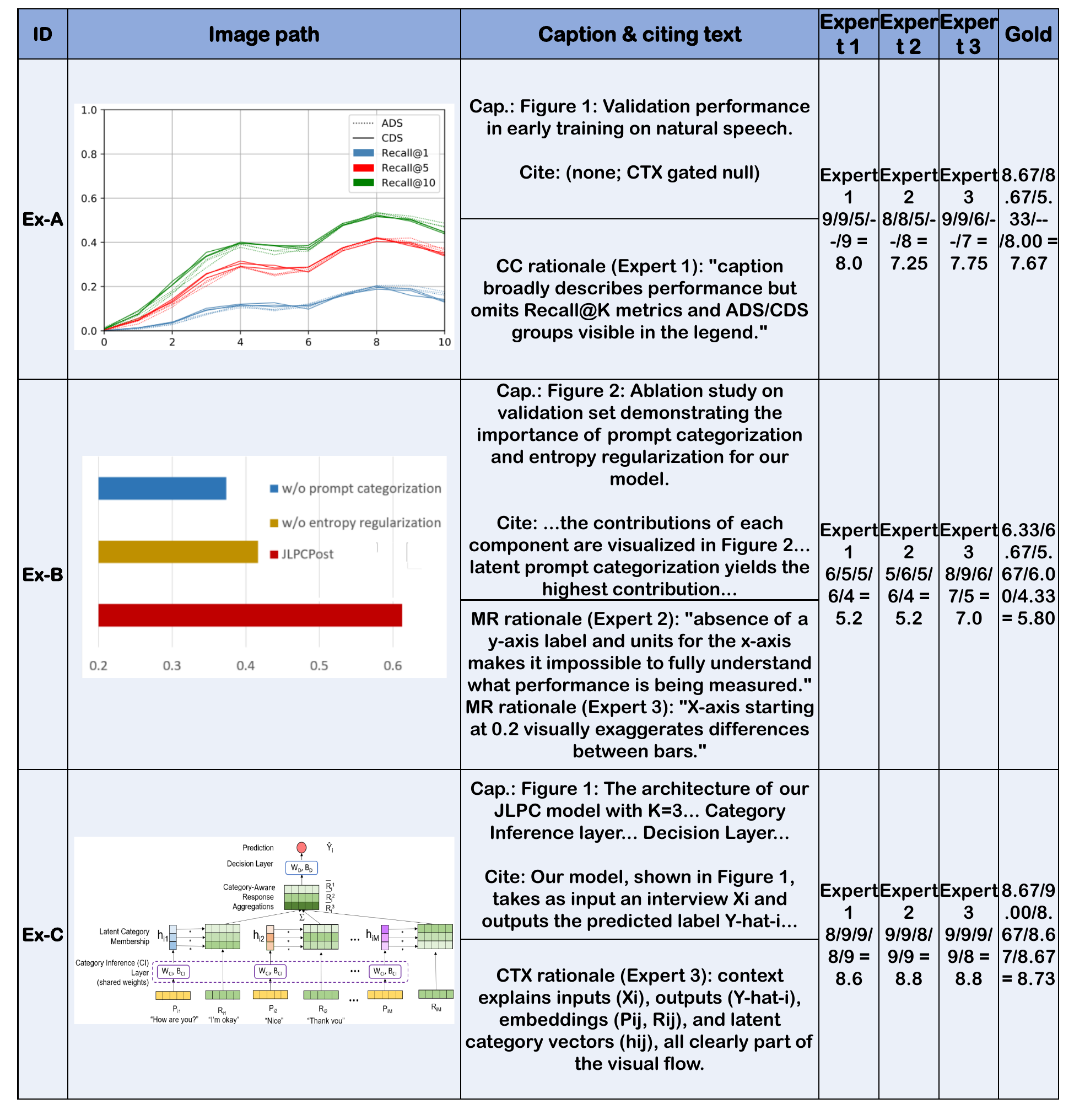}
\end{center}
\caption{Human multi-rater annotation cases from \textit{eval1200}.
Each block is one figure. Scores are on $[1,10]$ (higher is better; MR means lower misleading risk).
Gold column reports dimension-first means. \nscore{} marks gated-null CTX.
Rationale snippets are verbatim from annotator records.}
\label{fig:apx-human-cases}
\end{figure}

Figure~\ref{fig:apx-human-cases} shows three \textit{eval1200} instances chosen to illustrate gating, disagreement, and agreement.
Ex-A has a caption but no citing text, so CTX is null and difficulty concentrates on CC.
Ex-B lacks clear axis labels. Raters~1 and~2 are substantially harsher on MR than Rater~3 (per-rater overall gap $1.8$, below the holdout's two-point adjudication trigger).
Ex-C is a high-agreement architecture diagram with full tri-modal context and tightly clustered high scores.

\subsection{Human vs.\ Model Comparison Cases}
\label{app:model-cases}

\begin{figure}[ht]
\begin{center}
\includegraphics[width=\linewidth]{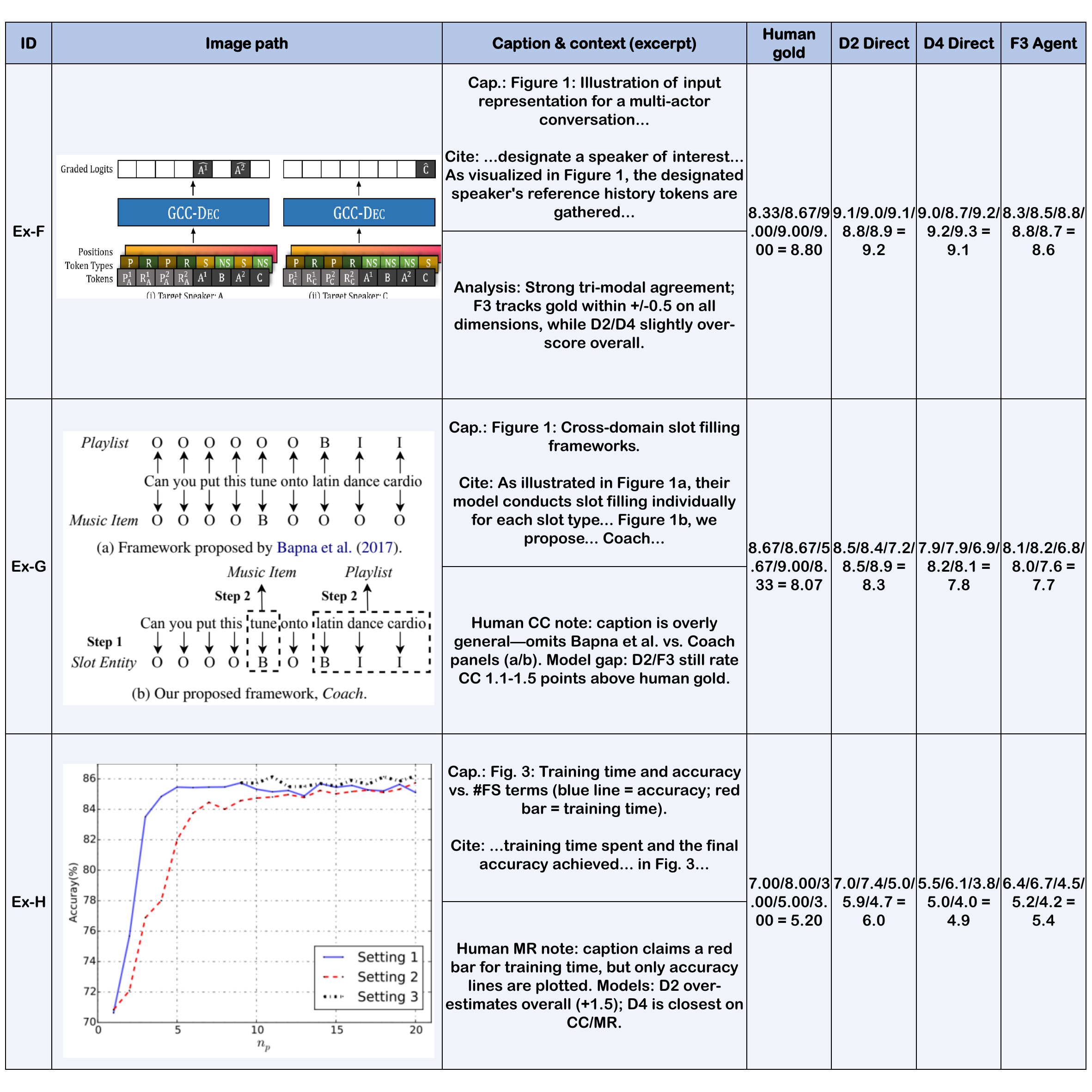}
\end{center}
\caption{Human gold vs.\ model predictions on selected \textit{eval1200} instances (disjoint from Figure~\ref{fig:apx-human-cases}).
Format: VC/SL/CC/CTX/MR $=$ Overall (unweighted mean of the five printed dimensions).
Human column is dimension-first gold. Model columns apply the same {evidence-availability gating} at inference. The Direct prompt requests integers in $[1,10]$; displayed cells may be one-decimal parses.}
\label{fig:apx-human-model}
\end{figure}

Figure~\ref{fig:apx-human-model} compares humans with Direct GPT-5.6-Sol (D2), Direct Qwen-VL-Max (D4), and SFQ-Agent GPT-5.6-Sol (F3) on three further instances.
Ex-F is already high-agreement. Staging changes little.
Ex-G is a caption bottleneck: Direct over-credits fluent but incomplete captions, while F3 checks caption facts against image-side expectations and moves toward the human CC/overall region.
Ex-H is a training-curve mismatch: the caption names a training-time bar that the plot does not draw, and staged judging moves CC toward the human gold by comparing the two evidence reports.
{The text stage does not see the image.
Its inputs on Ex-H are the released caption, which states that the blue line denotes accuracy and the red bar denotes training time, together with the citing paragraphs that discuss training time and final accuracy under different numbers of FS terms.
Its output is a language report with caption claims, a caption-implied image, and a text-claim severity written from the wording alone.
It does not output that the red bar is missing, and it does not emit the final cross-modal contradiction.
The vision stage reads only the image and the OCR/CV features.
On this figure the accuracy lines are readable and no training-time bar is drawn, so the vision report can set the intrinsic risk band to $[7,8]$ and omit the bar from the image facts.
The fusion call compares the two reports and only then creates the contradiction that the text requires a red bar while the image does not provide one, which it writes into CC and CTX.
To avoid reading the text-stage field as that missing-bar judgment, we rename it text-claim severity.
Algorithm~\ref{alg:runner} still computes MR from the visual band and this severity channel.
When the closed-form demonstration sets that channel to contradiction with no MR cap, a fusion draft at the visual midpoint would keep MR at $8$, and the Runner returns the fixed branch value $2$.
That $2$ is the rule function's fixed output on the contradiction branch for this single-branch walk-through, not an arbitrary lecture number and not the archived end-to-end score.
Figure~\ref{fig:apx-human-model} reports the stored F3 MR on Ex-H as $4.2$, which is the recorded Agent output for this figure under the same protocol.
}

\subsection{Annotation Interface and Adjudication}
\label{app:interface}

Each record stores five dimension scores, a per-rater gated overall, per-dimension reason strings, plus a summary and a suggestion.
Contested or low scores require reasons. Unfaithful rationales are rejected in review.
The full JSONL for all 1,200 \textit{eval1200} instances (with aggregated human means) is released at the project page and GitHub repository listed in the abstract.
On the ACL 2025 holdout, per-rater overall gaps above two points trigger adjudication before release.
Released gold then follows Eqs.~\ref{eq:dim-mean} and \ref{eq:gold-overall}.

\section{Extended analysis and limitations}
\label{app:extended-analysis}

Table~\ref{tab:overall-floor} makes the metric gap concrete: a constant-mean predictor already reaches W-1 $87.9\%$, so Within-1 barely separates protocols, while MAE and SRCC carry more of the signal. Even the highest observed SRCC is $0.598$, consistent with gold scores concentrated near $8$.

\begin{table}[ht]
\caption{Overall \textit{eval1200} floor versus protocol-best rows from Table~\ref{tab:protocol-ablation}. Model and constant W-1 are both versus \emph{mean} gold, not pairwise human agreement.}
\label{tab:overall-floor}
\centering
\scriptsize
\renewcommand{\arraystretch}{1.08}
\begin{tabularx}{0.82\linewidth}{@{}Xccc@{}}
\toprule
\rowcolor{tabheader}
\textbf{Predictor} & \textbf{MAE}$\downarrow$ & \textbf{W-1}$\uparrow$ & \textbf{SRCC}$\uparrow$ \\
\midrule
Constant mean ($8.13$) & 0.565 & 87.9\% & n/a \\
\rowcolor{tabalt}
Direct-best (Claude-Opus-4.8) & 0.443 & 92.4\% & 0.582 \\
Sidecar-best (GPT-5.6-Sol) & 0.436 & 92.6\% & 0.578 \\
\rowcolor{tabbest}
Agent-best (GPT-5.6-Sol) & 0.418 & 93.4\% & 0.598 \\
\bottomrule
\end{tabularx}
\end{table}

Backend Behavior.
GPT-5.6-Sol is the most balanced, with near-zero bias under Agent. Qwen-VL-Max is calibrated under Direct (W-1 88.8\%) but ranks poorly (SRCC 0.416). Staging recovers SRCC to 0.519 while CC MAE stays at $1.007$. GLM-4.6V improves most under richer protocols (Direct MAE $0.662$ to Agent MAE $0.574$), whereas Doubao stays negatively biased ($-0.120$ to $-0.098$). These are descriptive comparisons at unequal protocol-level call budgets.

Scope and Limitations.
Per-dimension error localizes failure along legibility, caption fit, narrative support, and misleading risk. The protocol ladder is a design choice, not a trained specialist judge, and CC remains a weak construct (pairwise W-1 $52.5\%$, $\alpha{=}0.49$ in Table~\ref{tab:human-ceiling}): fitting mean gold is easier than matching a rater, and published-only figures compress the score range. CTX is unavailable when no explicit citing paragraph is resolved. The evaluation uses bounded citing snippets rather than full-paper model context, the 29-run matrix is not a complete factorial, Agent uses more calls than Direct, and we do not report paper-clustered confidence intervals or repeated-API variance for the $0.025$ MAE gap. Rankings are therefore descriptive point estimates for frozen run IDs, not evidence of a universal ordering. Future releases should retain a paper-disjoint hidden test for leaderboard use and report uncertainty clustered by source paper.

\colorlet{blue}{black}
\section{Article-Level Figure Sets}
\label{app:article-cases}

Figure~\ref{fig:apx-human-cases} and Figure~\ref{fig:apx-human-model} each show one crop.
The five articles below are disjoint from those examples.
For each article we keep every released figure in \textit{eval1200}, and the indices run from 1 with no gaps.
All five dimensions are scored, so neither CC nor CTX is an L1 null.
Gold follows Eqs.~\ref{eq:dim-mean} and~\ref{eq:gold-overall}.
Scores lie on $[1,10]$.
Higher MR means lower misleading risk.
Source titles, venues, and years are omitted.
Raters are written as R1, R2, and R3 inside each case.
A per-rater vector is $(\mathrm{VC},\mathrm{SL},\mathrm{CC},\mathrm{CTX},\mathrm{MR})$.
These rows are human ratings. Model predictions are not attached.

The sets sit next to two results in the main text.
On \textit{eval1200}, CC is the weak axis (pairwise W-1 $52.5\%$, $\alpha{=}0.49$ in Table~\ref{tab:human-ceiling}, mean spread $2.65$ and median spread $3.0$ in Table~\ref{tab:apx-irr-stats}), while SL is tight (mean spread $1.03$).
Overall gold on \textit{eval1200} is $8.13$, and the articles below also land near that band.
A short caption can leave CC low even when the citing paragraph carries the claim, which is the caption and figure check under manuscript-linked context, read across the figures of one article.

\paragraph{Case 1. Three figures.}
\begin{table}[ht]
\caption{Case 1. Dimension scores and gated overall.}
\label{tab:apx-c1}
\begin{center}
\small
\setlength{\tabcolsep}{3.5pt}
\begin{tabular}{@{}llcccccc@{}}
\toprule
\rowcolor{tabheader}
\textbf{Fig.} & \textbf{Rater} & \textbf{VC} & \textbf{SL} & \textbf{CC} & \textbf{CTX} & \textbf{MR} & \textbf{Ov.} \\
\midrule
1 & R1 & 9 & 9 & 8 & 9 & 9 & 8.80 \\
\rowcolor{tabalt}
1 & R2 & 9 & 9 & 8 & 9 & 9 & 8.80 \\
1 & R3 & 8 & 9 & 5 & 7 & 7 & 7.20 \\
\rowcolor{tabalt}
1 & Gold & 8.67 & 9.00 & 7.00 & 8.33 & 8.33 & 8.27 \\
\midrule
2 & R1 & 7 & 8 & 8 & 9 & 9 & 8.20 \\
\rowcolor{tabalt}
2 & R2 & 9 & 9 & 6 & 8 & 9 & 8.20 \\
2 & R3 & 9 & 9 & 5 & 9 & 7 & 7.80 \\
\rowcolor{tabalt}
2 & Gold & 8.33 & 8.67 & 6.33 & 8.67 & 8.33 & 8.07 \\
\midrule
3 & R1 & 7 & 8 & 10 & 9 & 9 & 8.60 \\
\rowcolor{tabalt}
3 & R2 & 9 & 8 & 9 & 9 & 9 & 8.80 \\
3 & R3 & 9 & 9 & 10 & 10 & 6 & 8.80 \\
\rowcolor{tabalt}
3 & Gold & 8.33 & 8.33 & 9.67 & 9.33 & 8.00 & 8.73 \\
\bottomrule
\end{tabular}
\end{center}
\end{table}

Figure~1 is a block diagram.
A candidate document enters a news encoder on the left.
Clicked documents enter a user encoder on the right.
The two vectors meet at a dot product labeled as click probability.
The caption names the framework and does not mention the encoders or the predictor.
Figure~1 receives $(9,9,8,9,9)$, $(9,9,8,9,9)$, and $(8,9,5,7,7)$, so $\bar{s}=(8.67,9.00,7.00,8.33,8.33)$ and $\bar{y}=8.27$.
SL is $9$ for every rater.
CC is $8/8/5$.
The score of $5$ treats the caption as topically right but too thin for the drawn data flow.
The citing paragraph uses the figure to separate this matching setup from a later generation task, so CTX is $9/9/7$ and stays above CC.
Visual clarity is not the failure. The caption is.

Figure~2 is six histograms.
Panels (a) and (b) are title length and body length.
Panels (c) and (d) are entity counts in the title and in the body.
Panel (e) is topic share, and panel (f) is click-history length.
The topic labels in (e) are small, which is why one rater gives VC $7$.
The caption says the figure shows corpus and training-set statistics, and it does not name the six panels.
Figure~2 receives $(7,8,8,9,9)$, $(9,9,6,8,9)$, and $(9,9,5,9,7)$, so $\bar{s}=(8.33,8.67,6.33,8.67,8.33)$ and $\bar{y}=8.07$.
CC is $8/6/5$.
The low CC asks the caption to list the panels.
The citing paragraph assigns topics to panel (c) and click history to panel (f).
On the drawing, topics are panel (e) and click history is panel (f), so the panel letter for topics does not match.
Gold CTX is still $8.67$.
The same paragraph states a mean title length of $10.5$ and a mean body length of $549.0$, and both values fall inside the bins of panels (a) and (b).
CC gold $6.33$ against CTX gold $8.67$ is caption underspecification with partial citing-text support, inside one article.

Figure~3 is the generation diagram.
Three numbered arrows mark where a user vector enters the decoder.
The first initializes the decoder state, the second changes attention over the source words, and the third changes the choice between generating and copying.
The caption lists those three routes in the same order as the drawing.
Figure~3 receives $(7,8,10,9,9)$, $(9,8,9,9,9)$, and $(9,9,10,10,6)$, so $\bar{s}=(8.33,8.33,9.67,9.33,8.00)$ and $\bar{y}=8.73$.
CC is $10/9/10$, unlike Figures~1 and~2.
VC is $7$ for one rater because the numerals $1$, $2$, and $3$ are small and light.
One rater scores MR at $6$ because blocks such as $F_1$, $F_2$, and the copy probability are drawn without their internal computation.
The other two score MR at $9$.
Gold MR is $8.00$.
Inside this article, overall gold moves from $8.07$ to $8.73$, near the \textit{eval1200} mean of $8.13$, while CC moves from $6.33$ to $9.67$.

\begin{figure*}[t]
\centering
\includegraphics[width=0.58\textwidth]{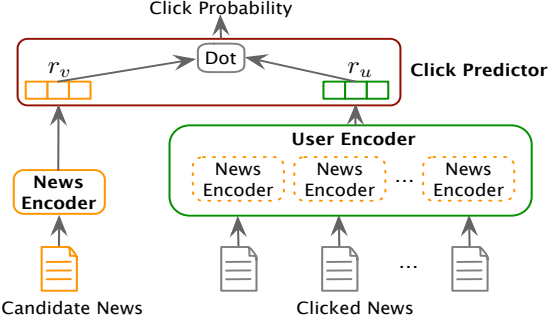}\\[0.35em]
{\small (a) Figure~1. $\bar{y}=8.27$.}\\[0.55em]
\begin{minipage}[t]{0.37\textwidth}
\centering
\includegraphics[width=\linewidth]{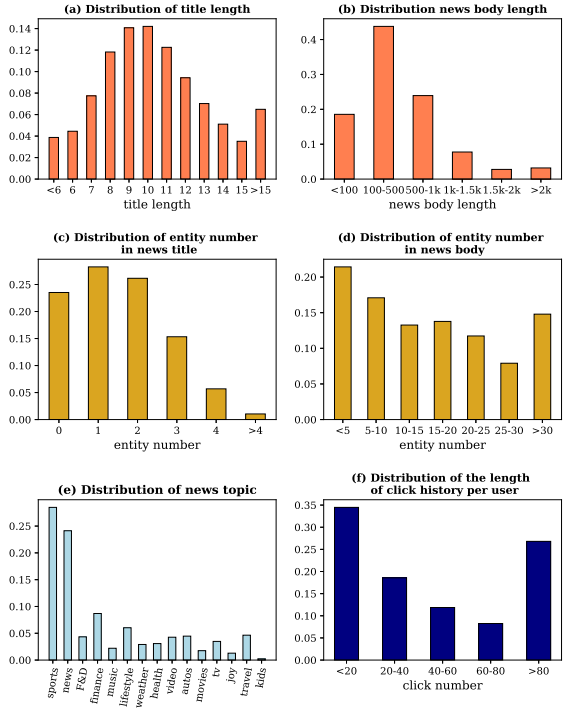}\\[0.25em]
{\small (b) Figure~2. $\bar{y}=8.07$.}
\end{minipage}\hfill
\begin{minipage}[t]{0.37\textwidth}
\centering
\includegraphics[width=\linewidth]{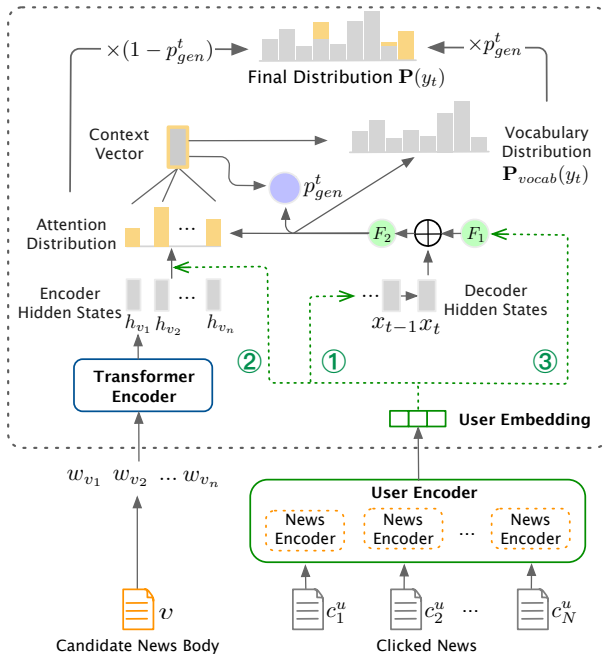}\\[0.25em]
{\small (c) Figure~3. $\bar{y}=8.73$.}
\end{minipage}
\caption{Case 1. Scores are in Table~\ref{tab:apx-c1}.}
\label{fig:apx-c1}
\end{figure*}

\paragraph{Case 2. Three figures.}
\begin{table}[ht]
\caption{Case 2. Dimension scores and gated overall.}
\label{tab:apx-c2}
\begin{center}
\small
\setlength{\tabcolsep}{3.5pt}
\begin{tabular}{@{}llcccccc@{}}
\toprule
\rowcolor{tabheader}
\textbf{Fig.} & \textbf{Rater} & \textbf{VC} & \textbf{SL} & \textbf{CC} & \textbf{CTX} & \textbf{MR} & \textbf{Ov.} \\
\midrule
1 & R1 & 9 & 9 & 10 & 9 & 9 & 9.20 \\
\rowcolor{tabalt}
1 & R2 & 9 & 9 & 7 & 9 & 9 & 8.60 \\
1 & R3 & 9 & 9 & 8 & 9 & 9 & 8.80 \\
\rowcolor{tabalt}
1 & Gold & 9.00 & 9.00 & 8.33 & 9.00 & 9.00 & 8.87 \\
\midrule
2 & R1 & 9 & 9 & 10 & 10 & 9 & 9.40 \\
\rowcolor{tabalt}
2 & R2 & 8 & 8 & 8 & 9 & 9 & 8.40 \\
2 & R3 & 8 & 9 & 6 & 9 & 8 & 8.00 \\
\rowcolor{tabalt}
2 & Gold & 8.33 & 8.67 & 8.00 & 9.33 & 8.67 & 8.60 \\
\midrule
3 & R1 & 8 & 9 & 10 & 10 & 8 & 9.00 \\
\rowcolor{tabalt}
3 & R2 & 8 & 8 & 6 & 7 & 8 & 7.40 \\
3 & R3 & 9 & 8 & 9 & 7 & 8 & 8.20 \\
\rowcolor{tabalt}
3 & Gold & 8.33 & 8.33 & 8.33 & 8.00 & 8.00 & 8.20 \\
\bottomrule
\end{tabular}
\end{center}
\end{table}

Figure~1 is a $2{\times}3$ grid of grouped bars, one panel per language code, with three series in each legend.
The vertical axis is the share of novel $n$-grams.
The caption states the metric, the contrast between neural outputs and human references, and that six datasets are shown.
It does not name the legend entries or the six language codes.
Figure~1 receives $(9,9,10,9,9)$, $(9,9,7,9,9)$, and $(9,9,8,9,9)$, so $\bar{s}=(9.00,9.00,8.33,9.00,9.00)$ and $\bar{y}=8.87$.
VC, SL, CTX, and MR are $9$ for every rater.
CC is $10/7/8$.
The score of $7$ is the rater who wants those names written into the caption.
The citing paragraph discusses the same novelty comparison, including one language where copying dominates, so CTX does not follow the CC dip.
SL spread is $0$, in line with the tight SL column in Table~\ref{tab:apx-irr-stats}.

Figure~2 is a scatter plot with five labeled points.
Both axes are titled on the canvas, and the caption defines the same two improvement rates.
Figure~2 receives $(9,9,10,10,9)$, $(8,8,8,9,9)$, and $(8,9,6,9,8)$, so $\bar{s}=(8.33,8.67,8.00,9.33,8.67)$ and $\bar{y}=8.60$.
CC is $10/8/6$.
The score of $6$ asks the caption to list the language codes already printed next to the points.
CTX is $10/9/9$.
The citing paragraph states the hypothesis the axes test, and it also says one language is left off the plot for scale.
That missing point is not a silent omission, which is why CTX stays high while CC can still drop.
MR is $9/9/8$.

Figure~3 has four stacked-area panels, one language in each corner, with year on the horizontal axis.
A shaded band marks 2019.
The caption names the four panel positions and the meaning of the shade.
Figure~3 receives $(8,9,10,10,8)$, $(8,8,6,7,8)$, and $(9,8,9,7,8)$, so $\bar{s}=(8.33,8.33,8.33,8.00,8.00)$ and $\bar{y}=8.20$.
CC is $10/6/9$.
The score of $6$ is because the topic names sit in the legends and not in the caption.
Some neighboring colors are close, so VC is $8/8/9$ rather than $9$ throughout, and MR is $8$ for every rater.
CTX is $10/7/7$.
Two raters note that the citing paragraph describes the shift over categories, and the six most frequent categories, but does not mention the shaded band.
The band is a caption fact, not a citing-text fact.
The three figures keep overall gold between $8.20$ and $8.87$, while single dimensions still move by several points.

\begin{figure*}[t]
\centering
\includegraphics[width=0.74\textwidth]{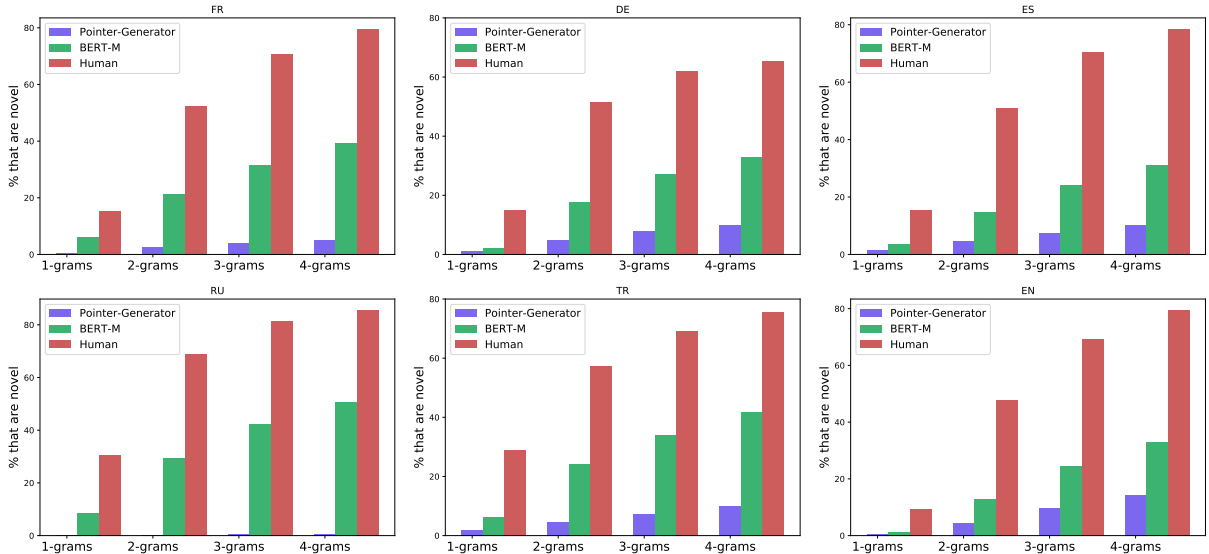}\\[0.3em]
{\small (a) Figure~1. $\bar{y}=8.87$.}\\[0.5em]
\includegraphics[width=0.42\textwidth]{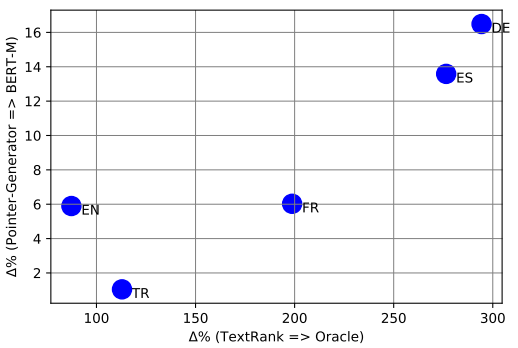}\\[0.3em]
{\small (b) Figure~2. $\bar{y}=8.60$.}\\[0.5em]
\includegraphics[width=0.74\textwidth]{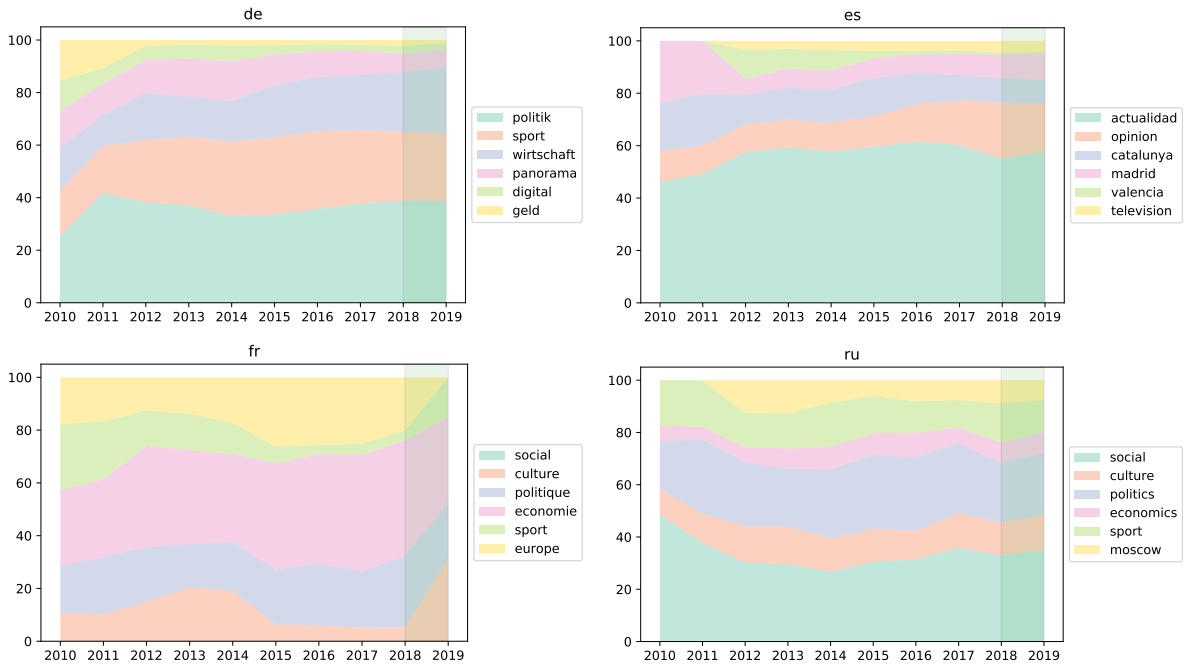}\\[0.3em]
{\small (c) Figure~3. $\bar{y}=8.20$.}
\caption{Case 2. Scores are in Table~\ref{tab:apx-c2}.}
\label{fig:apx-c2}
\end{figure*}

\paragraph{Case 3. Two figures.}
\begin{table}[ht]
\caption{Case 3. Dimension scores and gated overall.}
\label{tab:apx-c3}
\begin{center}
\small
\setlength{\tabcolsep}{3.5pt}
\begin{tabular}{@{}llcccccc@{}}
\toprule
\rowcolor{tabheader}
\textbf{Fig.} & \textbf{Rater} & \textbf{VC} & \textbf{SL} & \textbf{CC} & \textbf{CTX} & \textbf{MR} & \textbf{Ov.} \\
\midrule
1 & R1 & 9 & 7 & 6 & 7 & 7 & 7.20 \\
\rowcolor{tabalt}
1 & R2 & 8 & 8 & 7 & 9 & 8 & 8.00 \\
1 & R3 & 9 & 8 & 5 & 8 & 5 & 7.00 \\
\rowcolor{tabalt}
1 & Gold & 8.67 & 7.67 & 6.00 & 8.00 & 6.67 & 7.40 \\
\midrule
2 & R1 & 9 & 9 & 9 & 9 & 9 & 9.00 \\
\rowcolor{tabalt}
2 & R2 & 8 & 7 & 9 & 9 & 8 & 8.20 \\
2 & R3 & 9 & 9 & 9 & 10 & 8 & 9.00 \\
\rowcolor{tabalt}
2 & Gold & 8.67 & 8.33 & 9.00 & 9.33 & 8.33 & 8.73 \\
\bottomrule
\end{tabular}
\end{center}
\end{table}

Figure~1 has two stacked panels and a shared horizontal axis with no label.
Each panel has a left axis and a right axis, colored to match the two curves.
The upper pair compares an in-batch average with a development-set average.
The lower pair compares in-batch similarity with development-set similarity.
The lines are sharp, which is why VC gold is $8.67$.
The caption names sharpness of logits for in-batch sentences and the development set, and it does not say which panel is the average and which panel is the similarity.
Figure~1 receives $(9,7,6,7,7)$, $(8,8,7,9,8)$, and $(9,8,5,8,5)$, so $\bar{s}=(8.67,7.67,6.00,8.00,6.67)$ and $\bar{y}=7.40$.
CC is $6/7/5$.
MR is $7/8/5$.
The score of $5$ on MR comes from the unlabeled horizontal axis together with axis titles ($h_{,d}$ and similarity) that the caption never defines.
SL is $7/8/8$ because the panels have no subfigure letters.
CTX is $7/9/8$. The citing paragraph discusses logit variance and similarity, so it supports the figure more than the caption does.
This is the lowest overall gold in the five articles, and the drop is in CC and MR, not in VC.

Figure~2 has two panels.
The upper panel is a distillation-loss curve and the lower panel is a development score, each with a shuffle condition and a non-shuffle condition, plus a shaded band.
The upper panel has no horizontal-axis title. One rater gives SL $7$ for that.
The caption names both curves and both strategies.
Figure~2 receives $(9,9,9,9,9)$, $(8,7,9,9,8)$, and $(9,9,9,10,8)$, so $\bar{s}=(8.67,8.33,9.00,9.33,8.33)$ and $\bar{y}=8.73$.
CC is $9$ for every rater.
The citing paragraph describes the early drop of the non-shuffle loss and the continued drop under shuffling, which is what the two panels show, so CTX is $9/9/10$.
The two figures use a similar line style. Overall gold differs by $1.33$ because Figure~1 leaves the axes and the two metrics under-specified.

\begin{figure*}[t]
\centering
\includegraphics[width=0.58\textwidth]{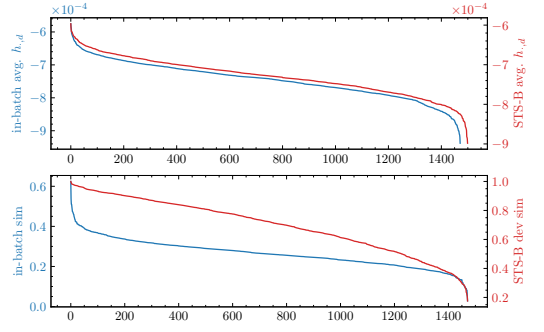}\\[0.3em]
{\small (a) Figure~1. $\bar{y}=7.40$.}\\[0.5em]
\includegraphics[width=0.58\textwidth]{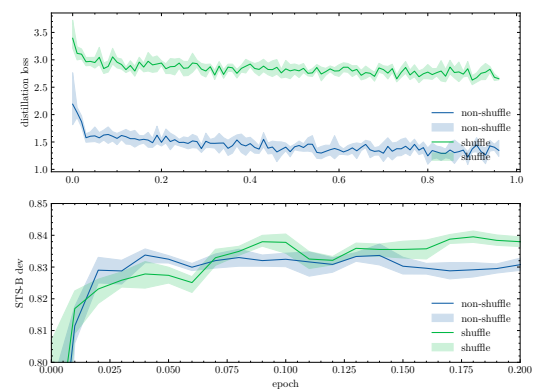}\\[0.3em]
{\small (b) Figure~2. $\bar{y}=8.73$.}
\caption{Case 3. Scores are in Table~\ref{tab:apx-c3}.}
\label{fig:apx-c3}
\end{figure*}

\paragraph{Case 4. Three figures.}
\begin{table}[ht]
\caption{Case 4. Dimension scores and gated overall.}
\label{tab:apx-c4}
\begin{center}
\small
\setlength{\tabcolsep}{3.5pt}
\begin{tabular}{@{}llcccccc@{}}
\toprule
\rowcolor{tabheader}
\textbf{Fig.} & \textbf{Rater} & \textbf{VC} & \textbf{SL} & \textbf{CC} & \textbf{CTX} & \textbf{MR} & \textbf{Ov.} \\
\midrule
1 & R1 & 9 & 9 & 9 & 9 & 9 & 9.00 \\
\rowcolor{tabalt}
1 & R2 & 9 & 9 & 9 & 9 & 9 & 9.00 \\
1 & R3 & 7 & 9 & 8 & 10 & 8 & 8.40 \\
\rowcolor{tabalt}
1 & Gold & 8.33 & 9.00 & 8.67 & 9.33 & 8.67 & 8.80 \\
\midrule
2 & R1 & 8 & 9 & 8 & 7 & 9 & 8.20 \\
\rowcolor{tabalt}
2 & R2 & 9 & 9 & 7 & 8 & 9 & 8.40 \\
2 & R3 & 9 & 9 & 9 & 6 & 6 & 7.80 \\
\rowcolor{tabalt}
2 & Gold & 8.67 & 9.00 & 8.00 & 7.00 & 8.00 & 8.13 \\
\midrule
3 & R1 & 9 & 9 & 10 & 10 & 9 & 9.40 \\
\rowcolor{tabalt}
3 & R2 & 8 & 9 & 9 & 9 & 9 & 8.80 \\
3 & R3 & 8 & 9 & 6 & 7 & 8 & 7.60 \\
\rowcolor{tabalt}
3 & Gold & 8.33 & 9.00 & 8.33 & 8.67 & 8.67 & 8.60 \\
\bottomrule
\end{tabular}
\end{center}
\end{table}

Figure~1 draws two examples side by side.
The left pair keeps a small substituent and changes the ring scaffold.
The right pair changes molecular size.
Labels and arrows are complete, and the drawing is not cut off at the border.
The caption states both examples, including which substructure is shared.
Figure~1 receives $(9,9,9,9,9)$, $(9,9,9,9,9)$, and $(7,9,8,10,8)$, so $\bar{s}=(8.33,9.00,8.67,9.33,8.67)$ and $\bar{y}=8.80$.
Two raters score every dimension at $9$.
The third lowers VC to $7$.
CTX gold is $9.33$, the highest dimension, because the citing paragraph walks through both examples in the same order as the drawing.

Figure~2 is a two-stage diagram.
The upper band is labeled environment inference and the lower band is labeled representation learning, with a decompose arrow into substructures, attentive pooling, and a predictor.
The caption describes that order and the two training losses.
Figure~2 receives $(8,9,8,7,9)$, $(9,9,7,8,9)$, and $(9,9,9,6,6)$, so $\bar{s}=(8.67,9.00,8.00,7.00,8.00)$ and $\bar{y}=8.13$.
SL is $9$ for every rater.
Some internal labels are small, so one VC score is $8$.
CTX is $7/8/6$ and is the lowest gold dimension on this figure.
The citing paragraph explains the substructure encoder and the attentive pooling in the lower band, and it does not walk through the upper band.
One rater also scores MR at $6$ for that gap.
High SL with lower CTX is the split the rubric is meant to keep. Layout quality does not stand in for citing-text support.

Figure~3 places two box plots beside a four-curve risk plot.
Panel letters (a), (b), and (c) are printed under the axes.
Panel (a) varies $k$, panel (b) varies $\beta$, and panel (c) shows risk against epoch for four environments.
The vertical axis of (a) and (b) is ROC-AUC.
Figure~3 receives $(9,9,10,10,9)$, $(8,9,9,9,9)$, and $(8,9,6,7,8)$, so $\bar{s}=(8.33,9.00,8.33,8.67,8.67)$ and $\bar{y}=8.60$.
SL is $9$ for every rater.
CC is $10/9/6$.
The score of $6$ is because the caption names $k$, $\beta$, and the risk curves, but does not name ROC-AUC.
CTX is $10/9/7$.
The citing paragraph discusses the trends in (a) and (c) in detail, and it mentions $\beta$ for (b) without describing that panel's shape.
Across the article, SL gold is $9.00$ on every figure, while CTX gold moves from $9.33$ to $7.00$ and back to $8.67$.

\begin{figure*}[t]
\centering
\includegraphics[width=0.70\textwidth]{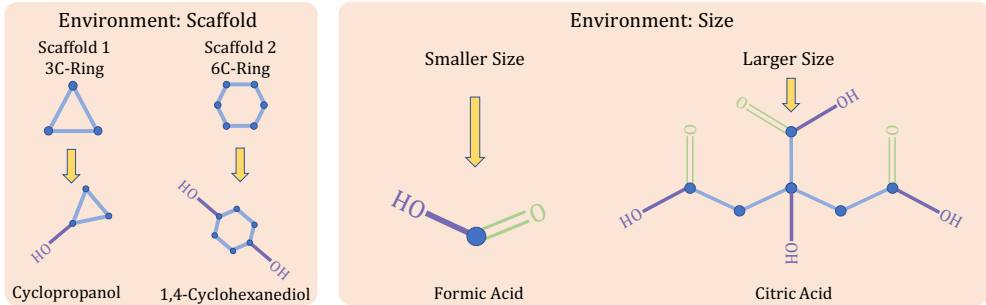}\\[0.3em]
{\small (a) Figure~1. $\bar{y}=8.80$.}\\[0.5em]
\includegraphics[width=0.70\textwidth]{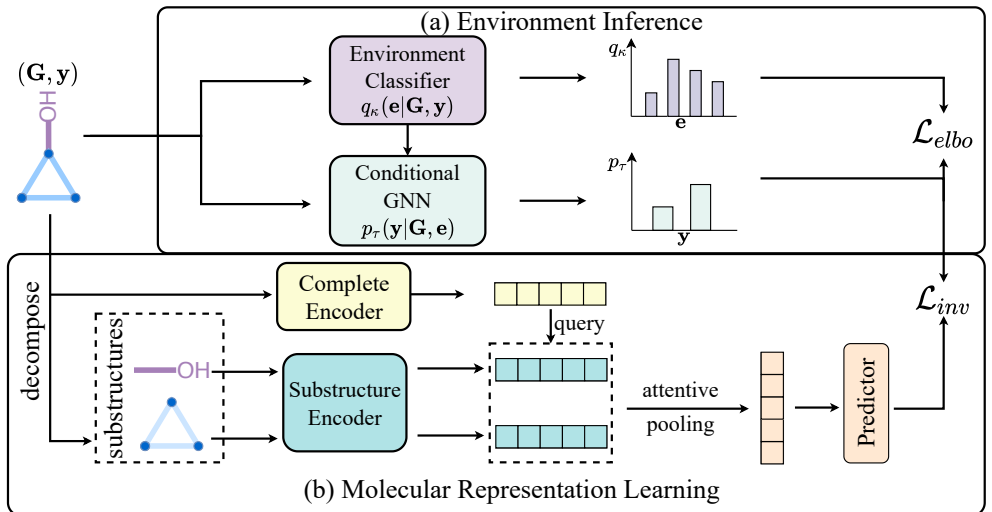}\\[0.3em]
{\small (b) Figure~2. $\bar{y}=8.13$.}\\[0.5em]
\includegraphics[width=0.74\textwidth]{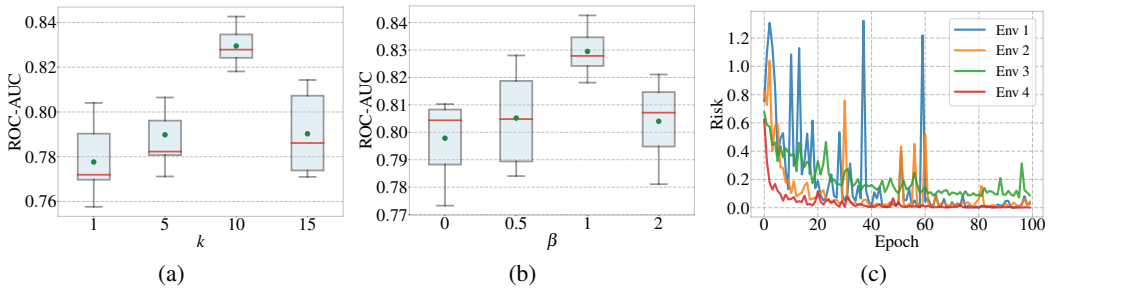}\\[0.3em]
{\small (c) Figure~3. $\bar{y}=8.60$.}
\caption{Case 4. Scores are in Table~\ref{tab:apx-c4}.}
\label{fig:apx-c4}
\end{figure*}

\paragraph{Case 5. Four figures.}
\begin{table}[ht]
\caption{Case 5. Dimension scores and gated overall.}
\label{tab:apx-c5}
\begin{center}
\small
\setlength{\tabcolsep}{3.5pt}
\begin{tabular}{@{}llcccccc@{}}
\toprule
\rowcolor{tabheader}
\textbf{Fig.} & \textbf{Rater} & \textbf{VC} & \textbf{SL} & \textbf{CC} & \textbf{CTX} & \textbf{MR} & \textbf{Ov.} \\
\midrule
1 & R1 & 8 & 9 & 8 & 8 & 9 & 8.40 \\
\rowcolor{tabalt}
1 & R2 & 8 & 9 & 8 & 9 & 9 & 8.60 \\
1 & R3 & 7 & 9 & 4 & 9 & 7 & 7.20 \\
\rowcolor{tabalt}
1 & Gold & 7.67 & 9.00 & 6.67 & 8.67 & 8.33 & 8.07 \\
\midrule
2 & R1 & 9 & 9 & 9 & 9 & 9 & 9.00 \\
\rowcolor{tabalt}
2 & R2 & 9 & 9 & 9 & 10 & 9 & 9.20 \\
2 & R3 & 9 & 9 & 4 & 7 & 9 & 7.60 \\
\rowcolor{tabalt}
2 & Gold & 9.00 & 9.00 & 7.33 & 8.67 & 9.00 & 8.60 \\
\midrule
3 & R1 & 7 & 9 & 8 & 9 & 9 & 8.40 \\
\rowcolor{tabalt}
3 & R2 & 9 & 9 & 7 & 8 & 8 & 8.20 \\
3 & R3 & 9 & 9 & 8 & 4 & 7 & 7.40 \\
\rowcolor{tabalt}
3 & Gold & 8.33 & 9.00 & 7.67 & 7.00 & 8.00 & 8.00 \\
\midrule
4 & R1 & 7 & 9 & 8 & 9 & 9 & 8.40 \\
\rowcolor{tabalt}
4 & R2 & 9 & 9 & 6 & 9 & 7 & 8.00 \\
4 & R3 & 9 & 9 & 7 & 7 & 7 & 7.80 \\
\rowcolor{tabalt}
4 & Gold & 8.33 & 9.00 & 7.00 & 8.33 & 7.67 & 8.07 \\
\bottomrule
\end{tabular}
\end{center}
\end{table}

Figures~1, 3, and~4 are attention heat maps.
Figure~2 is a grouped bar chart.
Token axes and color bars are present, and the panels are not cropped.
SL gold is $9.00$ on every figure, with every rater at $9$.
Several VC scores are $7$ because color-bar ticks and axis text are small.
Layout stays stable, and the disagreement sits in CC, CTX, and MR.
That repeats the \textit{eval1200} pattern in Table~\ref{tab:apx-irr-stats}.

Figure~1 shows two maps side by side, titled $\sqrt{64}$ and $\sqrt{96}$.
High weights form short diagonal bands, and on the right map those bands reach later token rows.
The caption calls the left map normal and the right map smoothed, and it does not tie those words to the two temperature titles.
Figure~1 receives $(8,9,8,8,9)$, $(8,9,8,9,9)$, and $(7,9,4,9,7)$, so $\bar{s}=(7.67,9.00,6.67,8.67,8.33)$ and $\bar{y}=8.07$.
CC is $8/8/4$.
The score of $4$ is that missing link between the caption wording and the panel titles.
The same rater gives MR $7$ for the same reason, and CTX $9$, because the citing paragraph describes shorter copy spans and attention that extends past the first $200$ tokens.
Gold CC is the dimension-first mean of $8$, $8$, and $4$, the same aggregation as the worked example in Section~\ref{app:aggregation}.
CC spread on this figure is $4$, above the \textit{eval1200} median CC spread of $3.0$.

Figure~2 plots the share of attention weights at or above $0.15$, in five bins of normalized token position.
The legend is $\lambda{=}1.0$, $1.5$, and $2.0$.
The caption says the comparison is across attention temperatures, and it does not mention $\lambda$.
Figure~2 receives $(9,9,9,9,9)$, $(9,9,9,10,9)$, and $(9,9,4,7,9)$, so $\bar{s}=(9.00,9.00,7.33,8.67,9.00)$ and $\bar{y}=8.60$.
VC, SL, and MR are $9$ for every rater.
CC is again $9/9/4$, from the temperature wording against the $\lambda$ legend.
CTX is $9/10/7$.
The citing paragraph defines the $0.15$ cutoff, the five bins, and the shift from the head of the document toward the tail, so two raters score CTX at $9$ and $10$ even though one of them scores CC at $4$.
Gold CC $7.33$ is $(9{+}9{+}4)/3$. One harsh caption reading moves the gold, and it does not move SL.

Figure~3 shows three maps in one row, at $\sqrt{64}$, $\sqrt{96}$, and $\sqrt{128}$.
The color bar runs higher than in Figure~1, up to about $0.6$, and a dark diagonal is visible in the left and middle panels.
Figure~3 receives $(7,9,8,9,9)$, $(9,9,7,8,8)$, and $(9,9,8,4,7)$, so $\bar{s}=(8.33,9.00,7.67,7.00,8.00)$ and $\bar{y}=8.00$.
CC is $8/7/8$.
The middle score wants the three temperature values written into the caption.
The split that matters is CTX, at $9/8/4$.
The citing paragraph discusses a shorter summary and a phrase taken from the end of the document.
It does not map those outputs onto $\sqrt{64}$, $\sqrt{96}$, and $\sqrt{128}$, and the heat maps show token indices rather than the words themselves.
That is why one rater scores CTX at $4$ and MR at $7$, while the other two still find the paragraph supportive.

Figure~4 is a second three-panel example in the same format.
The document axis runs past $700$ tokens, unlike Figure~3, whose axis stops near $200$.
The left and middle maps stay near the top of that axis. The right map also marks bands near $350$ and near $700$.
Figure~4 receives $(7,9,8,9,9)$, $(9,9,6,9,7)$, and $(9,9,7,7,7)$, so $\bar{s}=(8.33,9.00,7.00,8.33,7.67)$ and $\bar{y}=8.07$.
CC is $8/6/7$ because the caption calls this a second example and does not say that the document is long, which is the visible difference from Figure~3.
MR is $9/7/7$ for the same omission.
CTX is $9/9/7$.
The citing paragraph states that the document has more than $700$ words and that one setting stays near the first $150$ words while another also uses the middle and the tail.
Here the citing text carries the claim that the caption leaves out.
Overall gold on the four figures is $8.07$, $8.60$, $8.00$, and $8.07$.
The article stays near the \textit{eval1200} mean. The movement is inside CC, CTX, and MR, not in SL.

\begin{figure*}[t]
\centering
\begin{minipage}[t]{0.38\textwidth}
\centering
\includegraphics[width=\linewidth]{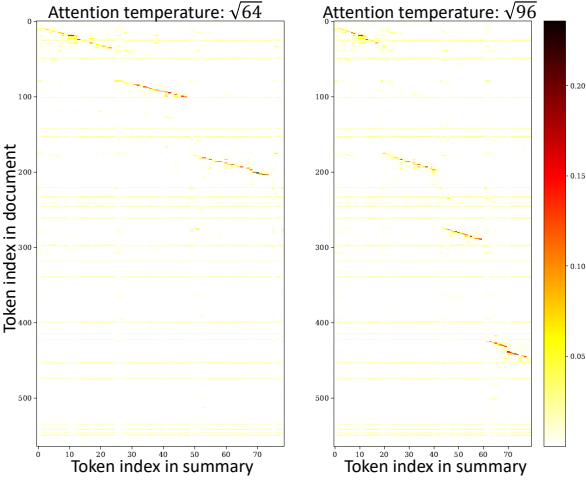}\\[0.25em]
{\small (a) Figure~1. $\bar{y}=8.07$.}
\end{minipage}\hfill
\begin{minipage}[t]{0.38\textwidth}
\centering
\includegraphics[width=\linewidth]{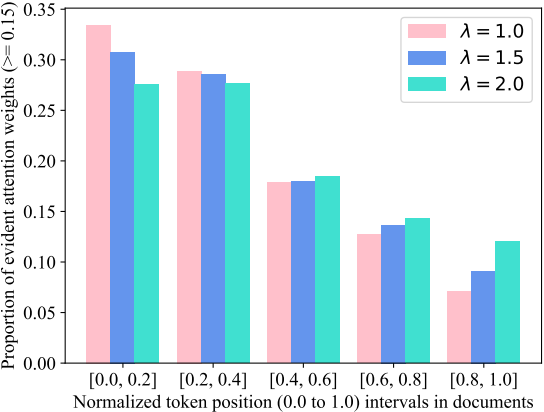}\\[0.25em]
{\small (b) Figure~2. $\bar{y}=8.60$.}
\end{minipage}\\[0.6em]
\includegraphics[width=0.74\textwidth]{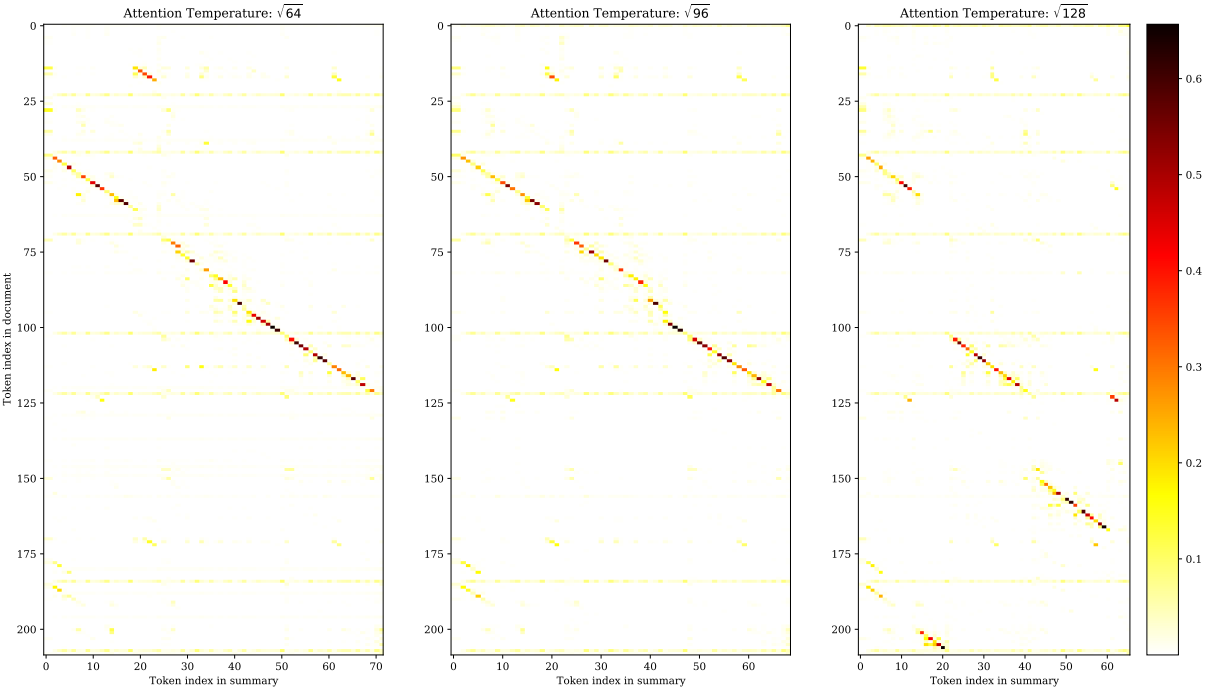}\\[0.3em]
{\small (c) Figure~3. $\bar{y}=8.00$.}\\[0.45em]
\includegraphics[width=0.74\textwidth]{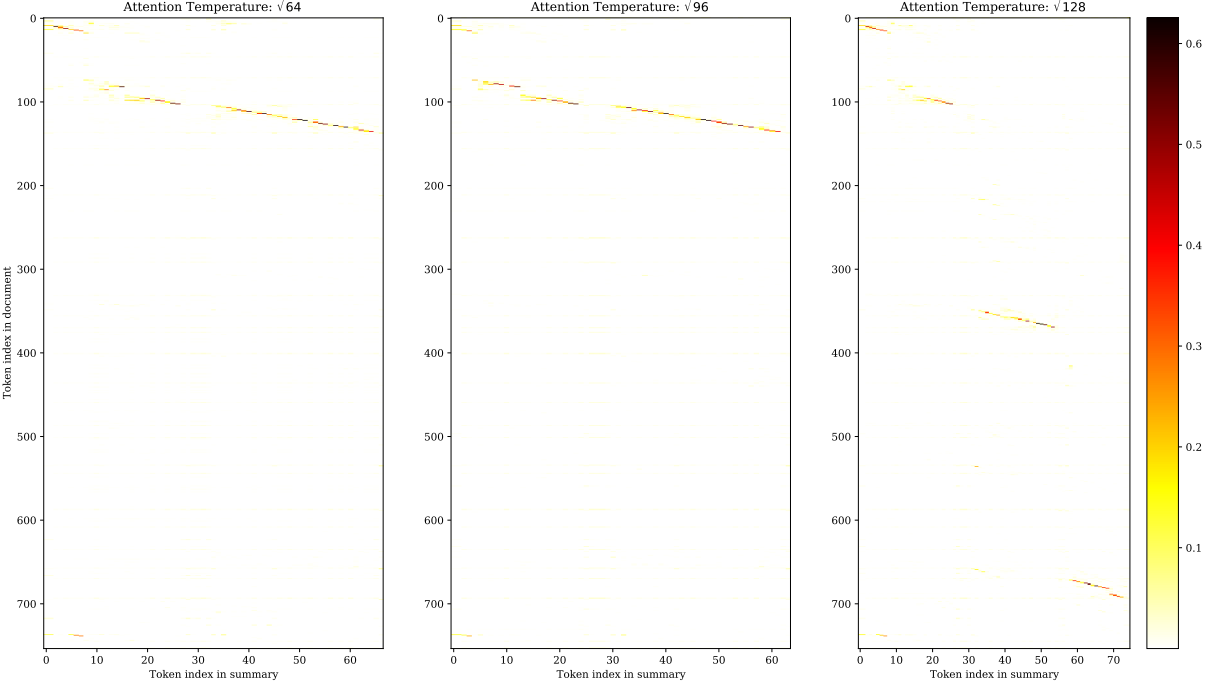}\\[0.3em]
{\small (d) Figure~4. $\bar{y}=8.07$.}
\caption{Case 5. Scores are in Table~\ref{tab:apx-c5}.}
\label{fig:apx-c5}
\end{figure*}

\colorlet{blue}{black}
{\section{Explanation faithfulness}
\label{app:faithfulness}

Table~\ref{tab:protocol-ablation} measures agreement with mean human scores on the $1{,}200$ figures in \textit{eval1200}.
Agreement with a score does not show that an explanation is grounded.
We drew $100$ of those figures with a fixed seed and split the Direct explanation (D2) and the Agent explanation (F3, the same backbone) into claims.
A claim is supported when the image, the caption, or the citing text states it.
A claim is unsupported when none of them states it.
A claim is contradicted when one of them states the opposite.
A sentence that correctly records a mismatch is supported.
Table~\ref{tab:apx-faith-sum} gives the counts.
On the $66$ figures whose crops are in the release, a figure contains an unsupported or contradicted claim for $45$ Direct explanations and for $5$ Agent explanations.
Figure~\ref{fig:apx-faith} prints three figures from the draw, and the tables under it give the claims.

\begin{table}[ht]
\caption{Claim labels on the $100$-figure draw. Each explanation is split into claims.}
\label{tab:apx-faith-sum}
\begin{center}
\small
\setlength{\tabcolsep}{5pt}
\begin{tabular}{@{}lccccc@{}}
\toprule
\rowcolor{tabheader}
\textbf{Protocol} & \textbf{Claims} & \textbf{Supported} & \textbf{Unsupported} & \textbf{Contradicted} & \textbf{Figures} \\
\midrule
Direct (D2) & 300 & 202 & 43 & 55 & 79/100 \\
\rowcolor{tabalt}
SFQ-Agent (F3) & 300 & 286 & 14 & 0 & 14/100 \\
\bottomrule
\end{tabular}
\end{center}
\end{table}

\begin{figure}[ht]
\begin{center}
\includegraphics[width=0.86\linewidth]{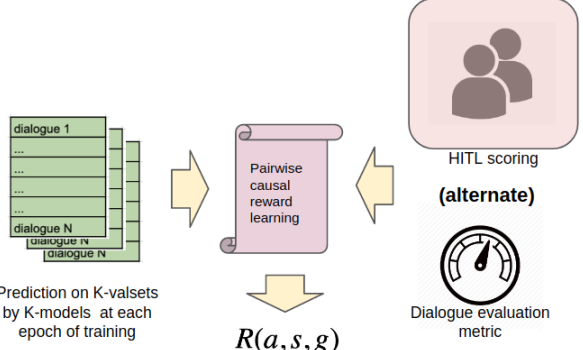}\\[0.35em]
{\small (a) A flowchart. The caption calls this a score display.}\\[0.55em]
\includegraphics[width=0.78\linewidth]{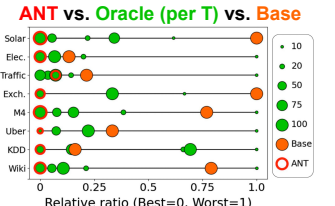}\\[0.35em]
{\small (b) ANT, Oracle, and Base. The caption names ANT and Oracle.}\\[0.55em]
\includegraphics[width=0.92\linewidth]{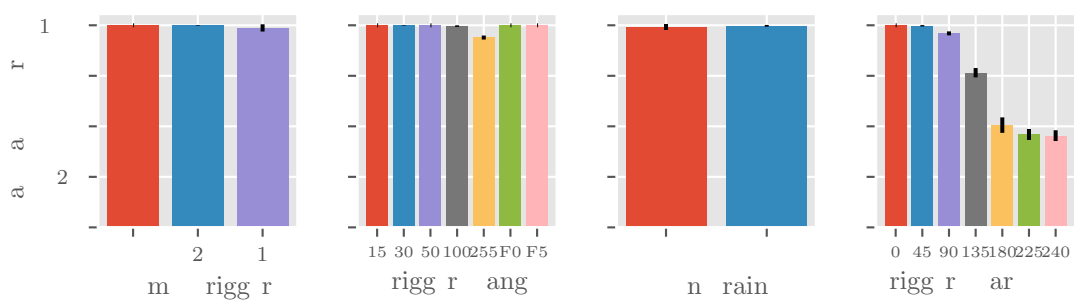}\\[0.35em]
{\small (c) Four bar-chart panels. The caption describes a 50-round accuracy curve.}
\end{center}
\caption{Three figures from the $100$-figure draw. The crops are the released images.}
\label{fig:apx-faith}
\end{figure}

\paragraph{Figure~\ref{fig:apx-faith}(a).}
The caption is ``Figure 6: Mixed Human-in-the-loop and automatic evaluation metric scores for pairwise causal reward learning.''
The released record has no citing text.
The canvas is a flowchart. Dialogue predictions enter pairwise causal reward learning, marked $R(a,s,g)$, and the flow then splits into HITL scoring, marked alternate, and a dialogue evaluation metric.
No numeric score is printed.

\begin{table}[ht]
\caption{Claims for Figure~\ref{fig:apx-faith}(a).}
\label{tab:apx-faith-a}
\begin{center}
\small
\setlength{\tabcolsep}{4pt}
\begin{tabular}{@{}p{0.07\linewidth}p{0.48\linewidth}p{0.16\linewidth}p{0.22\linewidth}@{}}
\toprule
\rowcolor{tabheader}
 & \textbf{Claim} & \textbf{Label} & \textbf{Ground} \\
\midrule
D & The figure shows metric scores. & Contradicted & The canvas shows no scores. \\
\rowcolor{tabalt}
D & Predictions go through $R(a,s,g)$ and then split into HITL scoring and a dialogue evaluation metric. & Supported & The blocks and the arrows are on the figure. \\
D & The caption matches the figure. & Contradicted & The caption says scores. The figure shows a process. \\
\rowcolor{tabalt}
A & The image is a flowchart and includes an alternate HITL branch. & Supported & The branch is marked alternate. \\
A & The canvas shows no numeric scores. & Supported & No score is printed. \\
\rowcolor{tabalt}
A & The caption calls the figure a score display, and the image is a process diagram. & Supported & The caption and the image differ in this way. \\
\bottomrule
\end{tabular}
\end{center}
\end{table}

\paragraph{Figure~\ref{fig:apx-faith}(b).}
The caption is ``Figure 10: ANT vs.\ Oracle.''
The citing sentence that names this figure says that ANT often outperforms the baseline TSDiff at smaller $T$ and reaches oracle performance in most cases.
The title printed on the chart is ``ANT vs.\ Oracle (per $T$) vs.\ Base''.
The vertical axis lists eight datasets.
The horizontal axis is relative ratio, with best at $0$ and worst at $1$.
The legend encodes bubble size as $10$, $20$, $50$, $75$, and $100$, and it gives separate colors to Base and ANT.

\begin{table}[ht]
\caption{Claims for Figure~\ref{fig:apx-faith}(b).}
\label{tab:apx-faith-b}
\begin{center}
\small
\setlength{\tabcolsep}{4pt}
\begin{tabular}{@{}p{0.07\linewidth}p{0.48\linewidth}p{0.16\linewidth}p{0.22\linewidth}@{}}
\toprule
\rowcolor{tabheader}
 & \textbf{Claim} & \textbf{Label} & \textbf{Ground} \\
\midrule
D & The figure compares ANT and Oracle. & Supported & Both names appear in the caption and on the chart. \\
\rowcolor{tabalt}
D & The caption states the comparison shown in the figure. & Contradicted & The chart title also includes Base and per $T$. \\
D & The caption covers the citing sentence about smaller $T$ and oracle performance. & Contradicted & The citing sentence states both. The caption states neither. \\
\rowcolor{tabalt}
A & The chart includes Base, eight datasets, a relative-ratio axis, and bubble sizes. & Supported & Each of these is printed on the figure. \\
A & The caption does not name Base, the ratio, the datasets, or the bubble sizes. & Supported & The caption and the figure, read together. \\
\rowcolor{tabalt}
A & The citing sentence states a gain at smaller $T$, and the caption does not. & Supported & The citing sentence and the caption. \\
\bottomrule
\end{tabular}
\end{center}
\end{table}

\paragraph{Figure~\ref{fig:apx-faith}(c).}
The caption is ``Figure 16: Local backdoor test accuracy of adversary client across 50 rounds. Error bars indicate one standard error.''
The released record has no citing text.
The figure is four bar-chart panels with categorical ticks.
The axis text is not readable as accuracy, as an adversary client, or as 50 rounds.
Some bars end in a cap.

\begin{table}[ht]
\caption{Claims for Figure~\ref{fig:apx-faith}(c).}
\label{tab:apx-faith-c}
\begin{center}
\small
\setlength{\tabcolsep}{4pt}
\begin{tabular}{@{}p{0.07\linewidth}p{0.48\linewidth}p{0.16\linewidth}p{0.22\linewidth}@{}}
\toprule
\rowcolor{tabheader}
 & \textbf{Claim} & \textbf{Label} & \textbf{Ground} \\
\midrule
D & The figure plots accuracy across 50 rounds. & Contradicted & The figure is four bar-chart panels. \\
\rowcolor{tabalt}
D & The plotted quantity is the local backdoor test accuracy of an adversary client. & Unsupported & The axis text is not readable. \\
D & The caps are error bars of one standard error. & Unsupported & Caps are visible. The figure does not define them. \\
\rowcolor{tabalt}
A & The figure is four bar-chart panels, not a line across 50 rounds. & Supported & The panels are visible. \\
A & The axis text is not readable as accuracy, adversary client, or 50 rounds. & Supported & Those words cannot be read on the axes. \\
\rowcolor{tabalt}
A & End caps are visible, and the figure does not define them as one standard error. & Supported & The caps are visible, and no definition is printed. \\
\bottomrule
\end{tabular}
\end{center}
\end{table}

On these three figures, Direct has $2$ supported claims, $5$ contradicted claims, and $2$ unsupported claims.
The Agent has $9$ supported claims.
}

\end{document}